\documentclass{article}

     \PassOptionsToPackage{square,sort,comma,numbers}{natbib}

     \usepackage[preprint]{neurips_2019}

\usepackage[utf8]{inputenc} 
\usepackage[T1]{fontenc}    
\usepackage{hyperref}       
\usepackage{url}            
\usepackage{booktabs}       
\usepackage{amsfonts}       
\usepackage{nicefrac}       
\usepackage{microtype}      

\usepackage{algorithm}
\usepackage{algorithmicx}
\usepackage{algpseudocode}
\usepackage{amsmath}
\usepackage{graphicx}

\usepackage{subfigure}
\usepackage{colortbl}
\definecolor{mygray}{gray}{.9}
\usepackage{comment}
\usepackage{amssymb}
\usepackage{wrapfig}

\title{Hierarchical Deep Multiagent Reinforcement Learning with Temporal Abstraction}

\author{%
Hongyao Tang\textsuperscript{1},
Jianye Hao\textsuperscript{1},
Tangjie Lv\textsuperscript{2},
Yingfeng Chen\textsuperscript{2},
Zongzhang Zhang\textsuperscript{3}, \\
{\bf
Hangtian Jia\textsuperscript{2},
Chunxu Ren\textsuperscript{2},
Yan Zheng\textsuperscript{1},
Zhaopeng Meng\textsuperscript{1},
Changjie Fan\textsuperscript{2},
Li Wang\textsuperscript{1}
} \\
  \textsuperscript{1}College of Intelligence and Computing, Tianjin University, \\
  \textsuperscript{2}Netease Fuxi AI Lab,
  \textsuperscript{3}Nanjing University \\
  \textsuperscript{1}\texttt{\{bluecontra,jianye.hao,yanzheng,mengzp,wangli\}@tju.edu.cn}, \\
  \textsuperscript{2}\texttt{\{hzlvtangjie,chenyingfeng1,jiahangtian,renchunxu,} \\
\texttt{fanchangjie\}@corp.netease.com},   \textsuperscript{3}\texttt{zhangzongzhang@gmail.com}
}

\begin{document}

\maketitle

\begin{abstract}
  Multiagent reinforcement learning (MARL) is commonly considered to suffer from non-stationary environments and exponentially increasing policy space.
  It would be even more challenging 
  when rewards are sparse and delayed over long trajectories.
  In this paper, we study hierarchical deep MARL 
  in cooperative multiagent problems with sparse and delayed reward.
  With temporal abstraction, we decompose the problem into a hierarchy of different time scales and investigate how agents can learn high-level coordination based on the independent skills learned at the low level.
  Three hierarchical deep MARL architectures
  are proposed to learn hierarchical policies under different MARL paradigms.
  Besides, we propose a new experience replay mechanism
  to alleviate the issue of the sparse transitions at the high level of abstraction and the non-stationarity of multiagent learning.
  We empirically demonstrate the effectiveness of our approaches in two domains with extremely sparse feedback:
  (1) a variety of Multiagent Trash Collection tasks,
  and (2) a challenging online mobile game, i.e., Fever Basketball Defense.
\end{abstract}

\section{Introduction}
Deep Reinforcement Learning (DRL) has
been applied to solve many challenging single-agent sequential decision-making problems in recent years
\cite{levine2016end,mnih2015human}.
However, many real-world tasks such as sensor networks \cite{zhang2013coordinating} and autonomous cars \cite{Cao2013AnOO}
are naturally organized as multiagent systems
(MASs),
where multiple agents have cooperative or competitive interactions with others.
In MASs, the problem complexity increases exponentially with the number of agents,
thus making it infeasible to apply traditional RL approaches directly.
Independent learning
alleviates this issue but usually fails to coordinate since it suffers from the non-stationary environments due to multiagent policy updates.

To address such challenges, a large amount of works have studied deep MARL from a variety of perspectives.
Foerster et al. \cite{foerster2017stabilising} propose multiagent importance sampling and fingerprints to stabilize experience replay of independent Q-leaning (IQL) \cite{tan1993multi}.
Peng et al. \cite{peng2017multiagent} propose multiagent Bidirectionally Coordinated Network (BiCNet) to learn effective collaborative strategies from a centralized perspective in StarCraft Micromanagement tasks.
Following the centralized training with decentralized execution paradigm \cite{Lowe2017MultiAgentAF},
Counterfactual multiagent policy gradients \cite{foerster2017counterfactual} are proposed
to address the multiagent credit assignment issue.
Besides, other approaches, e.g., CommNet \cite{sukhbaatar2016learning} and DIAL \cite{foerster2016learning},
are proposed to learn communication protocols among multiple cooperative agents.

Most previous works learn cooperative polices directly over primitive action spaces and usually perform well in environments with dense reward.
However, in many real-world scenarios
rewards are naturally sparse and immediate feedbacks are difficult to design manually.
In such circumstances, it is challenging to achieve coordination and most plain MARL approaches can hardly learn effective policies from sparse and delayed feedback.
To address this issue, one important direction is hierarchical MARL with temporal abstraction \cite{ghavamzadeh2006hierarchical},
which was firstly proposed in tabular RL settings and is still neglected in deep MARL literature.
One representative work in tabular settings is the multiagent MAXQ \cite{makar2001hierarchical}.
With temporal abstraction,
the difficulties of learning with sparse reward are reduced by manually decomposing the original problem into multiple levels of abstractions.
The results show that cooperative strategies can be efficiently learned at the high level based on the low-level skills,
in simple Multiagent Trash Collection tasks and Automated Guided Vehicles (AGVs) scheduling tasks.
However, the effectiveness of hierarchical MARL need to be further verified in more complex and practical problems with deep learning techniques.

In this paper,
we study hierarchical MARL with deep neural networks in cooperative multiagent problems with sparse reward.
We show that temporal abstraction can facilitate multiagent policy learning and leverage this to construct
three hierarchical deep MARL architectures for different MARL learning paradigms.
Besides, a new experience replay mechanism is introduced to further improve the coordination during the high-level policy learning.

Key contributions of this work are summarized as follows:

\begin{itemize}
    \item To the best of our knowledge, we are the first to study hierarchical deep MARL with temporal abstraction.
    We model the problems with a two-level hierarchy of abstraction, and study how cooperative policies and independent skills can be learned at different temporal scales together.
    To this end, we propose three hierarchical deep MARL architectures, i.e., hierarchical Independent Learner (h-IL), hierarchical Communication network (h-Comm) and hierarchical Qmix network (h-Qmix), for different MARL paradigms.

    \item We discuss two issues of hierarchical MARL, including the sparse transitions of high-level learning and the non-stationarity of multiagent policy updates.
    To alleviate the above issues, we introduce Augmented Concurrent Experience Replay (ACER) through augmenting high-level transitions with sub-transitions and
    conducting experience replay concurrently.

    \item We show the effectiveness of hierarchical deep MARL in a variety of extended classic Multiagent Trash Collection tasks. Moreover, our approaches achieve good performance in the challenging Fever Basketball Defense game, in which
    plain MARL approaches can hardly learn effective cooperative strategies due to sparse and delayed rewards.

\end{itemize}

\section{Background}

\paragraph{Markov Games}
In this paper, we consider a multiagent extension of Markov decision processes (MDPs) called (partially observable) Markov games \cite{littman1994markov}.
A Markov game for $N$ agents is defined by a set of states $\mathcal{S}$,
a set of primitive actions $\{\mathcal{A}^i\}_{i=0}^{N}$ and a set of observations $\{\mathcal{O}^i\}_{i=0}^{N}$ for each agent.
Each agent $i$ receives a private observation $o^i : \mathcal{S} \rightarrow \mathcal{O}^i$ and selects actions through its policy $\pi^i: \mathcal{O}^i \times \mathcal{A}^i \rightarrow [0,1]$, producing the next state according to the state transition function $\mathcal{P}: \mathcal{S} \times \mathcal{A}^1 \times \dots \times \mathcal{A}^N \times \mathcal{S} \rightarrow [0,1]$.
Each agent $i$ obtains extrinsic rewards $r^i: \mathcal{S} \times \mathcal{A}^1 \times \dots \times \mathcal{A}^N \rightarrow \mathbb{R}$ from the environment.
Each agent $i$ aims to maximize its own total expected return $\sum_{t=0}^{T} \gamma^{t}r^i_t$,
where $\gamma$ is a discount factor and $T$ is the time horizon.
Especially, a Markov game is fully-cooperative when all agents share a joint utility function, i.e., the reward function is identical for all agents.

\paragraph{Temporal Abstraction}

\begin{wrapfigure}{r}{0cm}
\begin{minipage}[t]{.45\linewidth}
\centering
\subfigure[]{
\label{figure:hierarchical_control}
\includegraphics[width=1.0\textwidth]{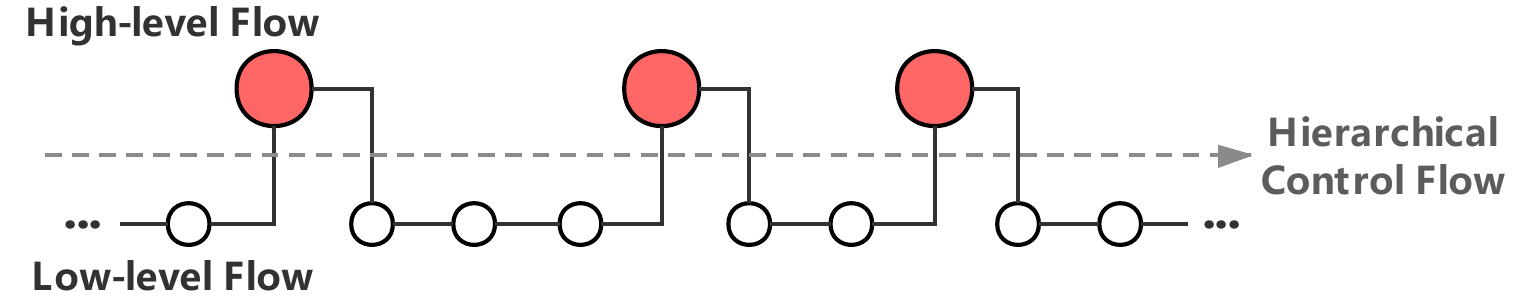}
}
\subfigure[]{
\label{figure:hierarhical_comparison}
\includegraphics[width=1.0\textwidth]{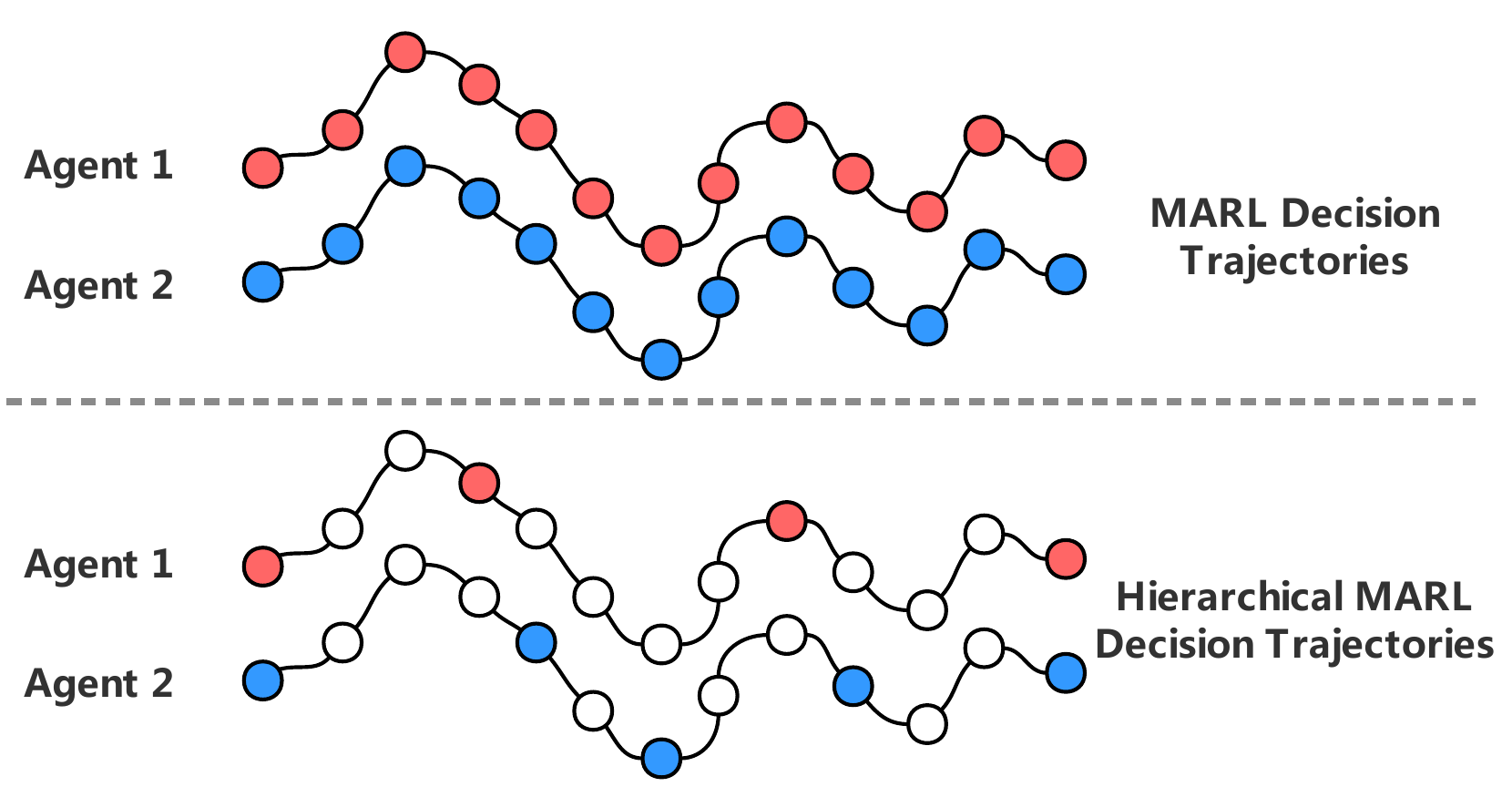}
}
\end{minipage}
\caption{
($a$) Hierarchical control flow. ($b$) Decision trajectories for MARL and hierarchical MARL.
For ($a$) and hierarchical MARL in ($b$), solid and hollow circles denote intrinsic goals and primitive actions respectively.
    }
\label{figure:2}
\end{wrapfigure}

Sparse and delayed feedback is a significant challenge in learning effective policies.
In order to tackle this, we introduce temporal abstraction.
Temporal abstraction is the key in solving long-period tasks,
since it reduces the difficulties of problem through allowing agents to ignore the details that are irrelevant for the task at hand
\cite{Fikes1972learning}.
This is commonly seen in the context of Hierarchical Reinforcement Learning (HRL) \cite{kulkarni2016hierarchical,Nachum2018Data,parr1998reinforcement},
where the tasks are usually decomposed into multi-level hierarchies
and the agent operates and learns simultaneously at multiple levels of abstraction.
Following the multiagent temporal abstraction literature \cite{ghavamzadeh2004learning,makar2001hierarchical,mehta2005multi},
we consider a two-level hierarchical model in a Markov game.
As shown in Figure \ref{figure:hierarchical_control},
each agent sets (multi-step) intrinsic goals at the high level and a sequence of primitive actions are made at the low level to achieve the certain goal.
We assume that the intrinsic goals are temporal abstraction of the simple tasks or skills that can be achieved or accomplished independently.
This reduces the difficulties of the original problem while maintains the multiagent properties,
thus agents can learn to cooperate and coordinate with others at the high level.

\paragraph{Hierarchical MARL}
With multiagent temporal abstraction, we introduce hierarchical MARL as illustrated in \ref{figure:hierarhical_comparison}.
The high level of hierarchy can be modeled as a Semi-Markov game, similar to the Multiagent Semi-MDP (MSMDP) \cite{ghavamzadeh2006hierarchical},
since intrinsic goals may last for multiple time steps.
Formally, each agent $i$ receives observation $o^i_t$ and chooses a goal $g_t^i \in \mathcal{G}^i$, where $\mathcal{G}^i$ denotes the set of all possible intrinsic goals.
A new goal $g_{t+\tau}^i$ is selected until the current goal $g_t^i$ is achieved or terminated after $\tau$ steps of low-level executions.
The next state $s_{t+\tau}$ is determined by the multi-step transition function $\tilde{\mathcal{P}}: \mathcal{S} \times \mathcal{A}^1 \times \dots \times \mathcal{A}^N \times \mathbb{N} \times \mathcal{S} \rightarrow [0,1]$, where $\mathbb{N}$ is the set of natural numbers.
The objective of agent $i$'s high-level policy $\pi^i$ is to maximize the cumulative extrinsic reward.

In contrast to the high level, we model the low level of hierarchy as MDPs.
Agent $i$ receives the intrinsic observation $\phi^i_t$
which depends on observation $o^i_t$ and current goal $g^i_t$,
and chooses an action $a_t^i \in \mathcal{A}^i_g$, where $\mathcal{A}^i_g$ ($\subseteq \mathcal{A}^i$) is a set of all available primitive actions under the current goal.
The intrinsic observation can be viewed as a form of state abstraction \cite{ghavamzadeh2006hierarchical} which excludes the state features that are irrelevant with the current goal.
Given an intrinsic goal, agent $i$ learns a low-level policy $\pi_g^i$ via optimizing the cumulative intrinsic reward.
For example, the intrinsic reward $\hat{r}^i_t$ can be $1$ if the goal is reached and $0$ otherwise \cite{kulkarni2016hierarchical}.

\subsection{Termination Models}
It should be noted that
the intrinsic goals of multiple agents may not terminate at the same time, thus inducing two termination models in hierarchical MARL, i.e., synchronous and asynchronous models \cite{ghavamzadeh2006hierarchical}.
In synchronous models,
agents make decisions at the same time points and this can be seen in many scenarios where synchronization mechanisms or conventions exist.
In such cases, it is suitable for high-level communication and centralized learning if possible.
However, synchronous models
may result in policy sub-optimality
since agents have to wait or early terminate their low-level executions to meet the synchronization requirements \cite{ghavamzadeh2006hierarchical}.
This can be even severe with the increase of agents.
In contrast, asynchronous termination models need no extra synchronization and decision making can take place in a fully decentralized fashion.
This allows agents to learn more flexible policies while may cause potential difficulties in achieving coordination.

\begin{figure}
\centering
\hspace{-0.5cm}
\subfigure[]{
\label{figure:h-IL}
\includegraphics[width=0.32\textwidth]{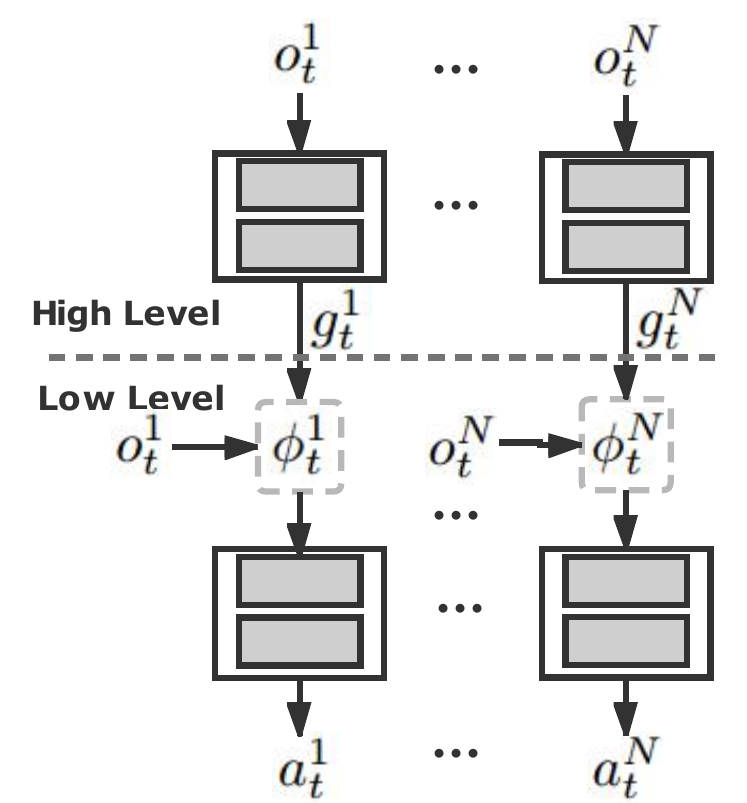}
\hspace{1.1cm}
}
\hspace{-1.7cm}
\subfigure[]{
\label{figure:h-Comm}
\includegraphics[width=0.36\textwidth]{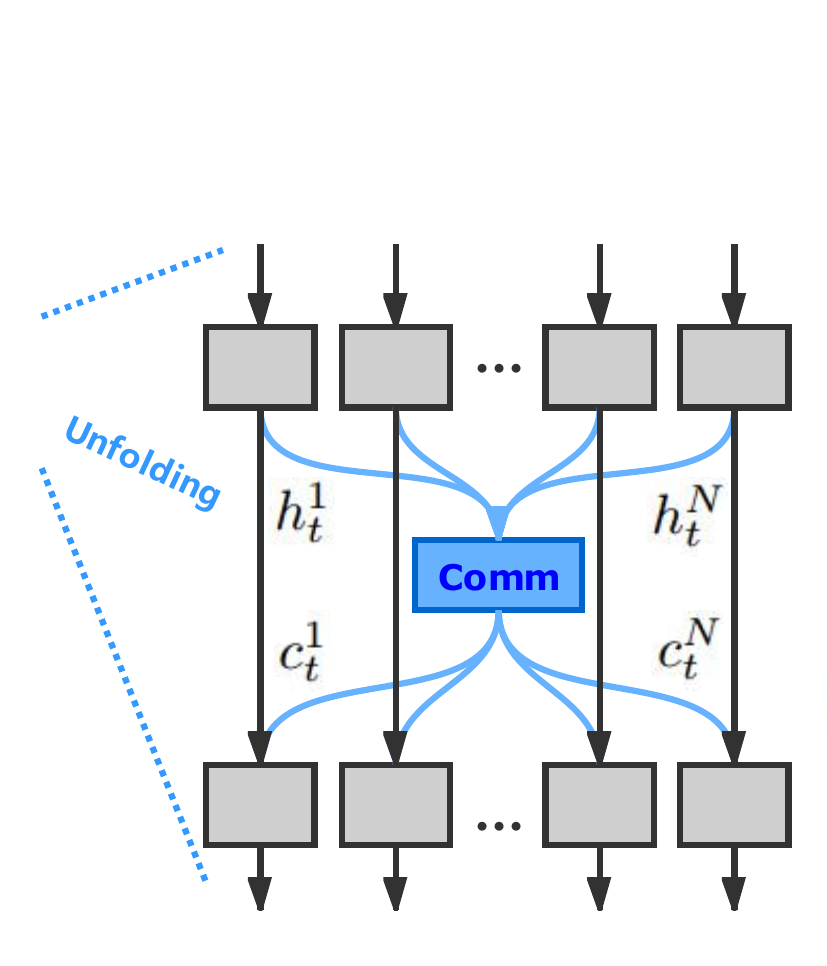}
\hspace{1.2cm}
}
\hspace{-0.6cm}
\subfigure[]{
\label{figure:h-Qmix}
\hspace{-1.2cm}
\includegraphics[width=0.29\textwidth]{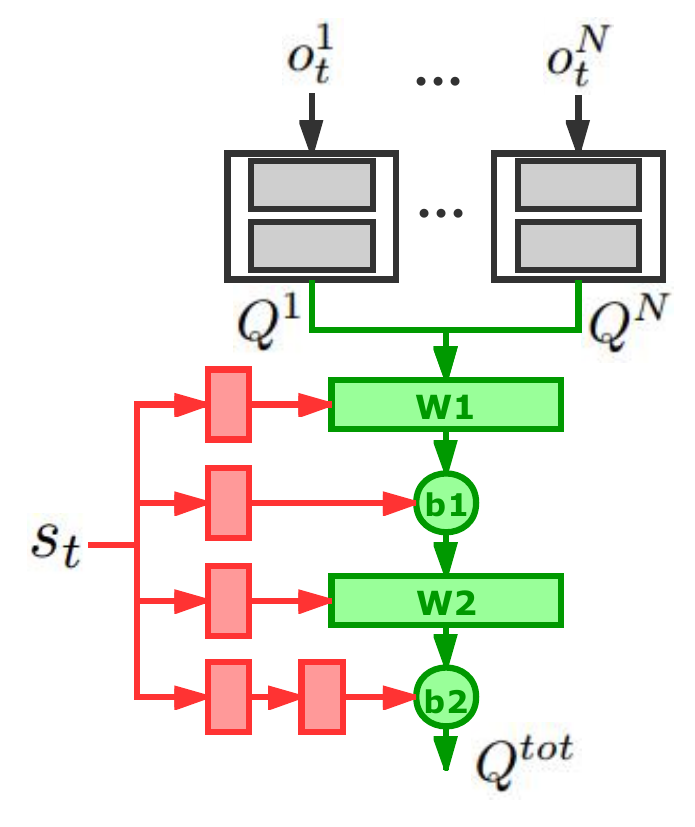}
}

\caption{Hierarchical deep MARL architectures. Grey rectangles denote the hidden layers of neural networks.
$(a)$ The h-IL architecture.
$(b)$ The h-Comm architecture.
$(c)$ The h-Qmix architecture, including the mixing network (green) and hypernetworks (red).
}
\label{figure:4}
\end{figure}

\section{Approaches}
In this section, based on the above modeling, we propose three architectures for hierarchical deep MARL
and a new experience replay mechanism to facilitate high-level learning.

\subsection{Hierarchical Deep MARL Architectures}

\paragraph{Hierarchical Independent Learner}
The most straightforward way is to let each agent learn its hierarchical policies independently, inducing the hierarchical Independent Learner (h-IL).
As illustrated in Figure \ref{figure:h-IL},
each h-IL agent updates its hierarchical policies independently and no gradient is propagated between high-level and low-level policies.
h-IL
can be implemented with either value-based or policy-based algorithms \cite{LillicrapHPHETS15continuous,SchulmanWDRK17proximal,van2016deep}.
In our experiments, we implement the h-IL architecture with DQNs \cite{mnih2015human}.
With Semi-MDP Q-Learning \cite{sutton1999between}, we can train the high-level policy $\pi$ parameterized by $\theta$ via minimizing the loss with mini-batch samples from replay buffer $\mathcal{D}$,
\begin{equation}
\label{equation:1}
  \mathcal{L}(\theta) = \mathbb{E}_{\left(s_t,g_t,\tau,r_{t:t+\tau-1}, s_{t+\tau}\right)  \sim \mathcal{D}} \left[ \left( y_t - Q(s_t,g_t;\theta) \right)^2 \right], \\
\end{equation}
where $y_t = \sum_{k=0}^{\tau-1}\gamma^{k}r_{t+k} + \gamma^{\tau}\max_{g_{t+\tau}} Q(s_{t+\tau},g_{t+\tau};\theta^{-})$.
Here $\theta^{-}$ denotes the parameters of the target value function and superscripts of agents are omitted for clarity.

The h-IL architecture is compatible with both synchronous and asynchronous termination models.
In independent learning paradigm, each agent only cares about its local information and treats other agents as part of the environment.
However,
it does not make use of available global information, which can be significant for facilitating coordination among cooperative agents.
To this end, we use h-IL as the base architecture and we derive the following two architectures from h-IL.

\paragraph{Hierarchical Communication Network}
In the scenarios where each agent can have access to other agents' information occasionally,
we propose hierarchical Communication network (h-Comm)
which allows agents to learn sparse high-level communication to facilitate cooperation.
Inspired by commNet \cite{sukhbaatar2016learning},
which was originally designed to learn temporal communication along consecutive steps,
h-Comm learns spatial communication between hidden layers among multiple agents.
As illustrated in Figure \ref{figure:h-Comm},
agent $i$ participates in the communication by feeding its hidden state $h^i_t$ into the Comm module and takes the processed output $c_t^i$ as additional inputs of the next layer.
One simple way to implement the \emph{Comm} module is to average over the hidden states of other agents, i.e.,
\begin{equation}
\label{equation:2}
c_t^i = \frac{ 1 }{ N-1 } \sum_{k \ne i} h_t^k.
\end{equation}
Intuitively, this allows each agent to explicitly reason about the information of others when choosing an intrinsic goal.
Note that h-Comm naturally requires agents to learn and execute high-level policies centrally.
However, it is not necessary for agents to choose goals in synchronization and thus h-Comm is also compatible with both synchronous and asynchronous termination models.

\paragraph{Hierarchical Qmix Network}
Since centralized control during online execution is usually difficult to maintain in practice, another popular way to facilitate coordination is to learn decentralized policies with centralized training.
Following this paradigm,
we propose hierarchical Qmix network (h-Qmix) which leverages the idea of the Qmix architecture \cite{rashid2018qmix} to coordinate the high-level policy updates during training.
As illustrated in Figure \ref{figure:h-Qmix},
a feedforward mixing network $\mathcal{M}$ (green) takes high-level action values $\{Q^i\}_{i=0}^{N}$ as inputs and mixes them monotonically, producing the joint value $Q^{tot}$:
\begin{equation}
\label{equation:qmix}
\begin{aligned}
	Q^{tot}(\vec{o}_t, \Vec{g}_t) = & \ \mathcal{M}\left( Q^1(o^1_t,g^1_t),\dots, Q^N(o^N_t,g^N_t) \right), \\
	& \frac{\partial Q^{tot}}{\partial Q^i} \ge 0, \forall i \in N,
\end{aligned}
\end{equation}
where $\Vec{o}_t$, $\Vec{g}_t$ are the joint observation and intrinsic goal of agents.
This ensures that a global \emph{argmax} operation performed on $Q^{tot}$ yields the same result as a set of individual \emph{argmax}  performed on individual $Q^i$.
The weights and biases of mixing network are produced by separate hypernetworks \cite{ha2017hypernetworks} (the red part of Figure \ref{figure:h-Qmix}), which takes state $s_t$ as input.
To enforce the monotonicity constraint in Equation \ref{equation:qmix},
the weights are restricted to be non-negative with an absolute activation function.


The h-Qmix architecture allows agents to train their high-level policies via updating a joint action-value function.
This is naturally compatible for synchronous termination model since $Q^{tot}$ is estimated over joint intrinsic goals.
For asynchronous case, an additional trim may be necessary to align the high-level transitions.
An additional discussion can be found in the Supplementary Material.

\paragraph{Low-level Parameter Sharing}
One advantage of our hierarchical model is that
low-level policies are relatively independent
and can be reused among agents.
Therefore,
we use parameter sharing in low-level learning from two aspects:
1)
we share parameters across the low-level policy networks of an agent that have similar input and output formats,
e.g., shooting and passing a football both use the move and kick actions;
2)
low-level policy parameters are shared among cooperative agents.
For specialization, the low-level policy takes a goal index and an agent identifier as additional inputs.

\begin{wrapfigure}{r}{0cm}
\begin{minipage}[t]{.4\linewidth}
\centering
\subfigure[]{
\label{figure:EA}
\includegraphics[width=1.0\textwidth]{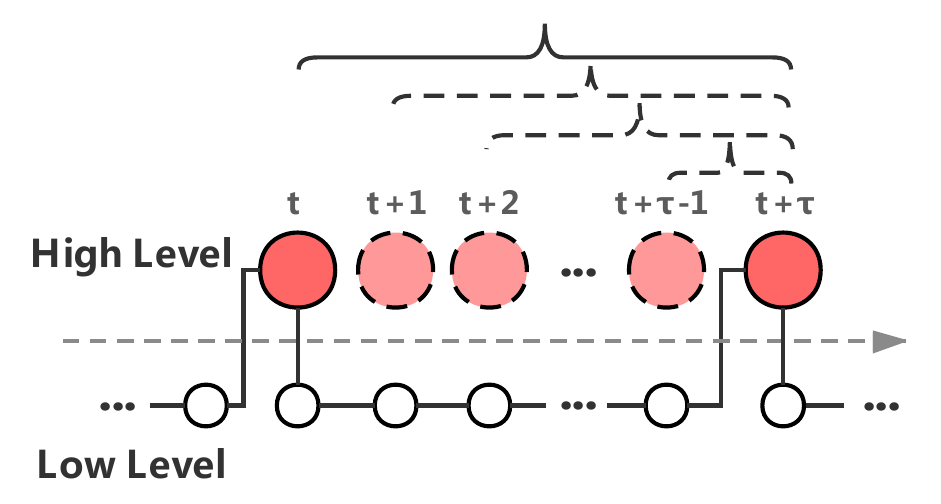}
}
\subfigure[]{
\label{figure:CR}
\includegraphics[width=0.9\textwidth]{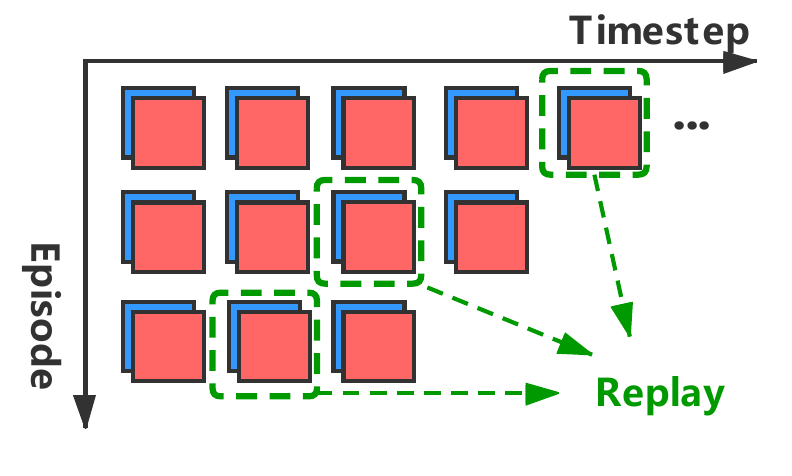}
}
\end{minipage}
\caption{
Illustrations for ACER.
$(a)$ Experience augmentation.
$(b)$ Concurrent sampling. Red and blue squares denote the experiences of different agents.
}
\label{figure:ACER}
\end{wrapfigure}

\subsection{Augmented Concurrent Experience Replay}
Experience replay is important for off-policy learning to improve the crucial sample efficiency.
However, in hierarchical MARL, experience replay can be flawed in two aspects.
First,
the high-level transitions start and terminate at sparse time points.
It causes inefficient updates of high-level policy since the intermediate states (sub-transitions) are not utilized.
Second,
high-level experiences can be obsolete and misleading
due to the non-stationary environment when replayed independently.
To address the above two issues,
we propose an experience replay mechanism named Augmented Concurrent Experience Replay (ACER), consisting of two components: experience augmentation and concurrent sampling.

As illustrated in Figure \ref{figure:EA},
we augmented the high-level transition $\left\langle o^i_t, g^i_t, r^i_{t:t+\tau-1}, o^i_{t+\tau} \right\rangle$ with sub-transitions (dashed braces), producing an augmented experience
$\left \{ \langle o^i_{t+k}, g^i_t, r^i_{t+k:t+\tau-1}, o^i_{t+\tau} \right\rangle\}_{k=0}^{\tau-1}$.
This enables agents to learn high-level policies more efficiently from denser experiences,
since each state encountered between consecutive goals can be updated.
In practice, to reduce the storage burden,
we can conduct experience augmentation with selective sub-transitions, e.g., interpolate at intervals.

Besides, to deal with the non-stationarity of high-level policy learning, we adopt the idea of concurrent sampling \cite{omidshafiei2017deep} in ACER.
As demonstrated in Figure \ref{figure:CR}, in each episode,
augmented experiences of agents
are stored in sequence along the timestep axis,
which ensures concurrent experiences are stored at the same position in the buffer.
Instead of sampling experiences independently,
ACER samples piles of concurrent experiences for all agents.
Concurrent sampling induces correlations in policy updates of agents, thus encouraging agents towards coordinated policies and stabilizing the training process as well.


\section{Experiments}
In this section, we firstly demonstrate the effectiveness of hierarchical deep MARL in extended Multiagent Trash Collection tasks,
and then evaluate our approaches in the challenging team sports game, i.e., Fever Basketball Defense.

\subsection{Multiagent Trash Collection (MATC)}

We devise a range of MATC tasks,
by extending the classic game in \cite{makar2001hierarchical}.
In all three tasks of Figure \ref{figure:MATC}, two agents (blue and red) are assigned to collect the cans (green)
and dump them into trash bins (yellow) in a grid-world with impenetrable walls (grey).
The observation of each agent is a image-like tensor of multiple channels that record the views
for different objects (i.e., agents, cans, trash bins and walls) in the environments.
Each agent has 7 actions, including navigation actions (i.e., \textbf{up}, \textbf{down}, \textbf{left}, \textbf{right} and \textbf{no-op}) and operation actions (i.e., \textbf{pick-up} and \textbf{put-down}).
For MATC-Room and MATC-Ring, agents obtain a reward of $0.5$ when a can is dumped into a trash bin.
For MATC-Coordination,
agents receive a reward of $1/0.5$ if both of them dump the can into the same (upper/lower) trash bin.
Otherwise, they obtain a reward of $0.1$ for dumping into different trash bins.
The three tasks have a max horizon of $100/100/50$ steps respectively.

\begin{figure}
\centering
\hspace{-0.0cm}
\subfigure[MATC Tasks]{
\label{figure:MATC}
\begin{minipage}{0.33\textwidth}
\includegraphics[width=0.45\textwidth]{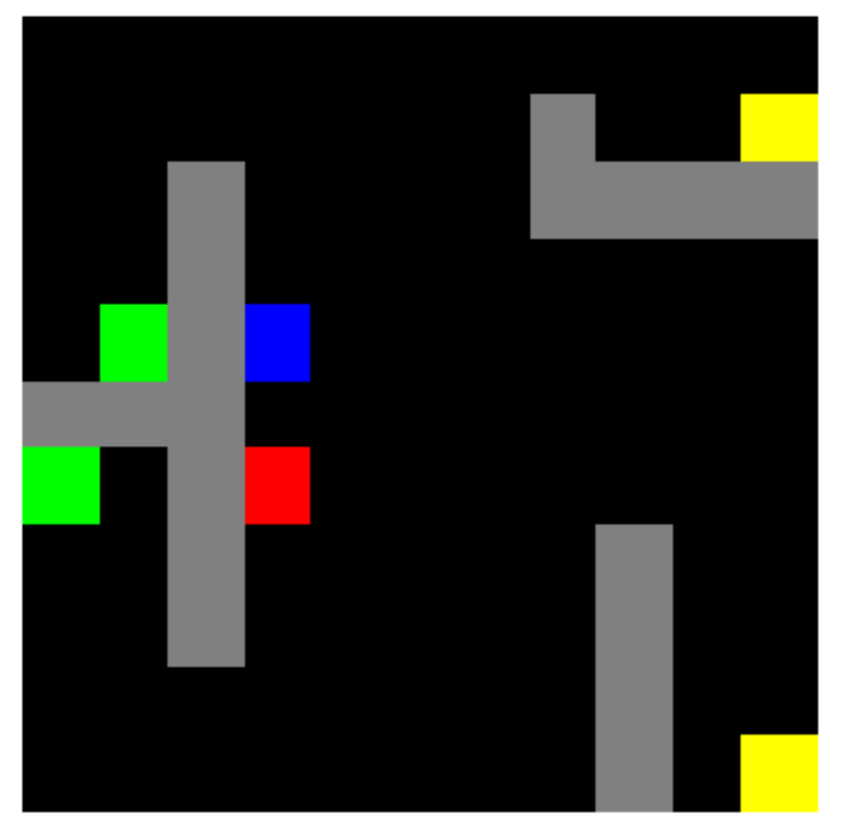}
\includegraphics[width=0.444\textwidth]{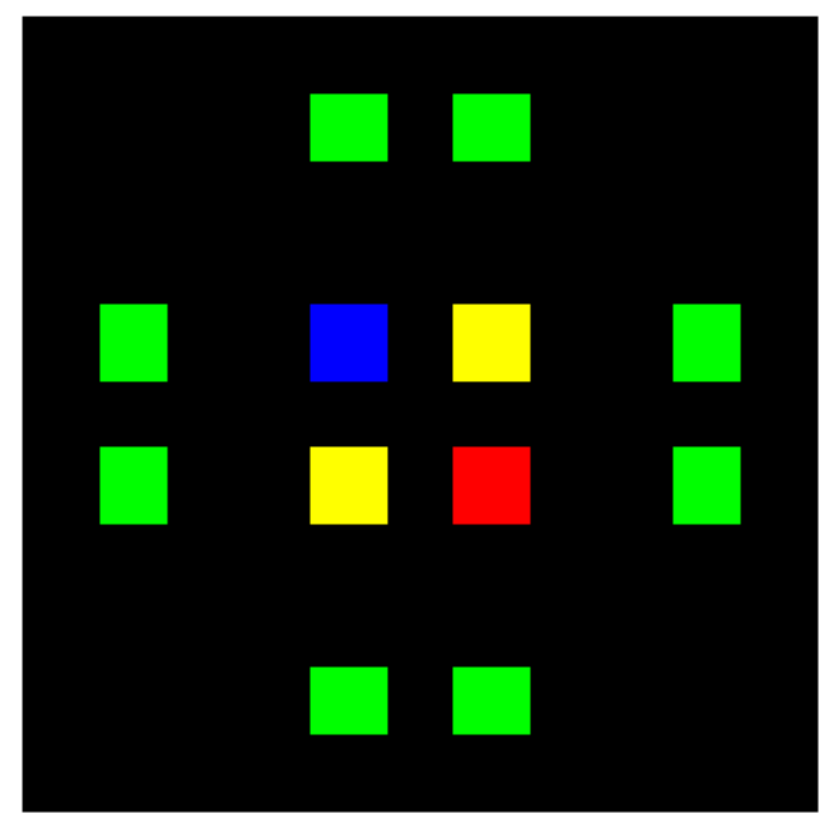} \\
\includegraphics[width=0.91\textwidth]{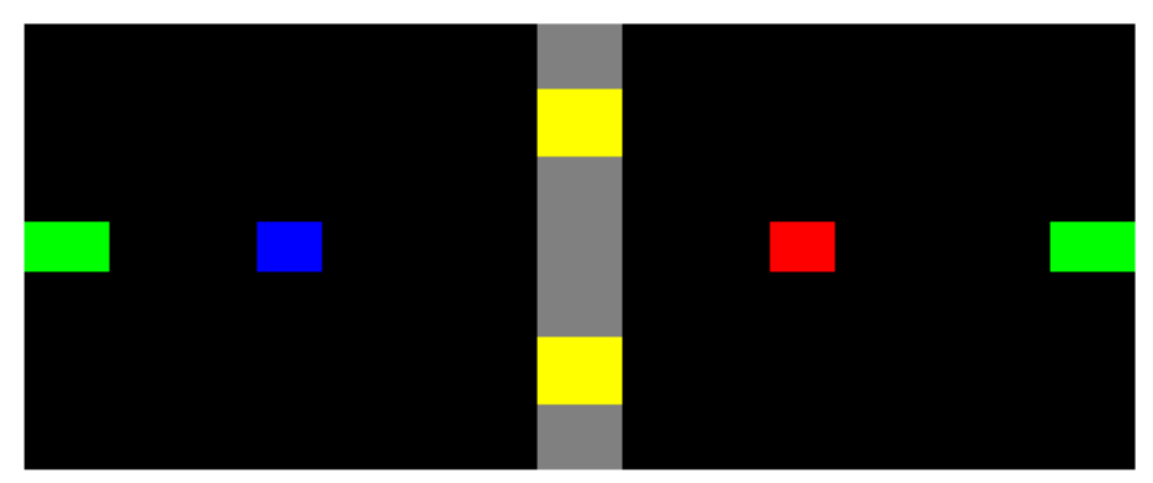}
\end{minipage}
}
\hspace{-0.6cm}
\subfigure[MATC-Room]{
\label{figure:room_results}
\begin{minipage}{0.22\textwidth}
\includegraphics[width=1\textwidth]{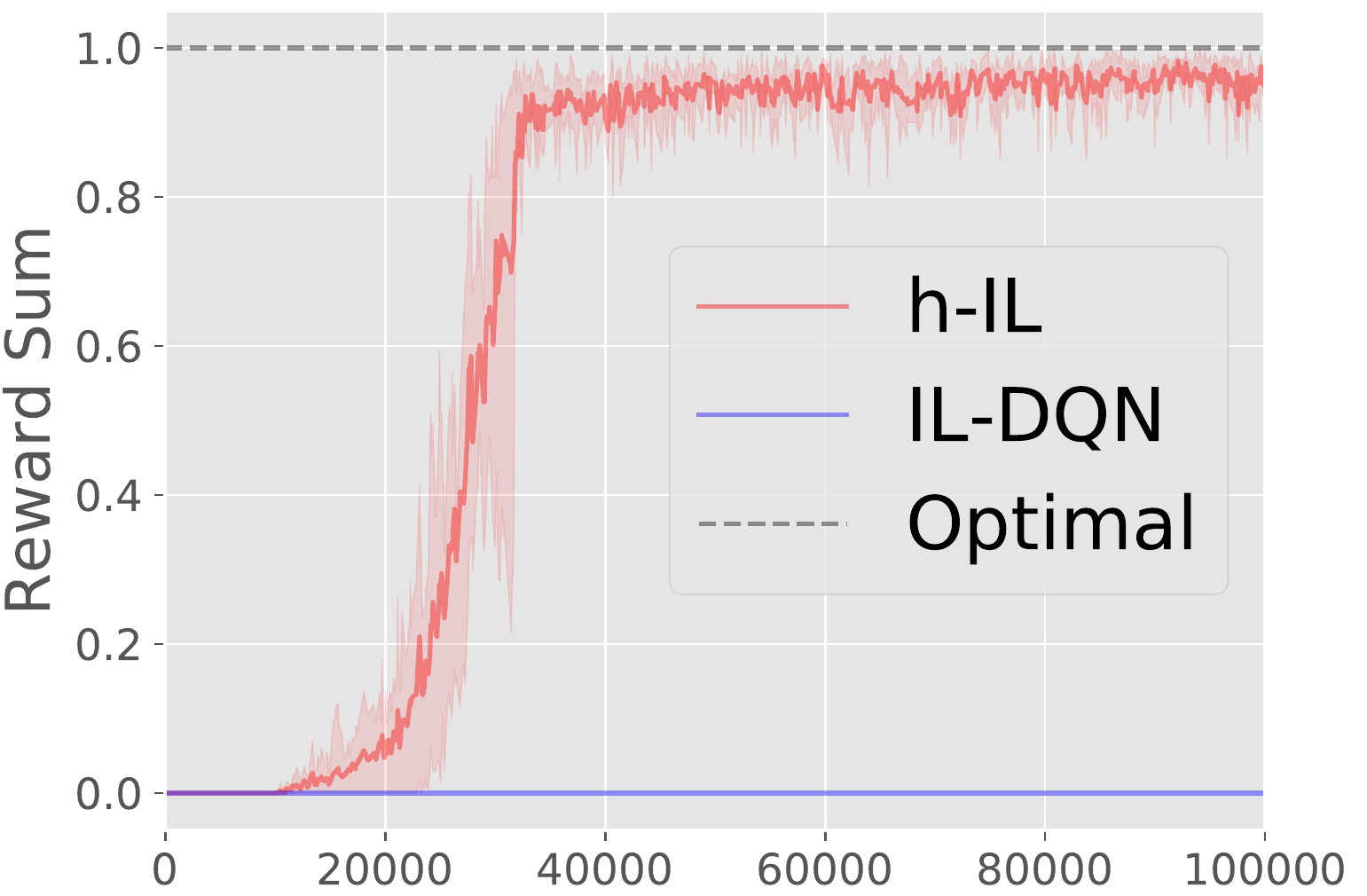} \\
\includegraphics[width=1\textwidth]{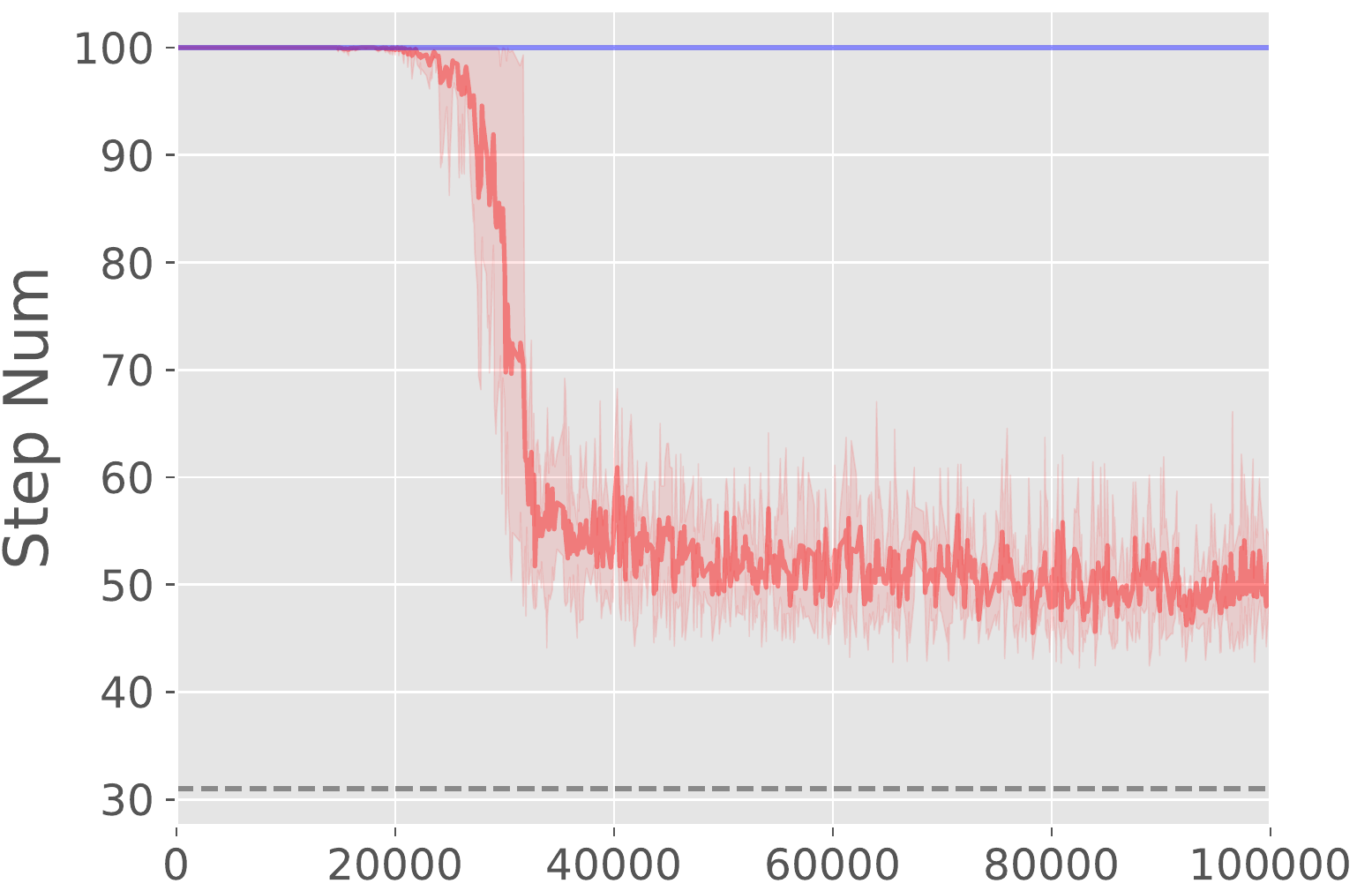}
\end{minipage}
}
\hspace{-0.2cm}
\subfigure[MATC-Ring]{
\label{figure:ring_results}
\begin{minipage}{0.21\textwidth}
\includegraphics[width=1\textwidth]{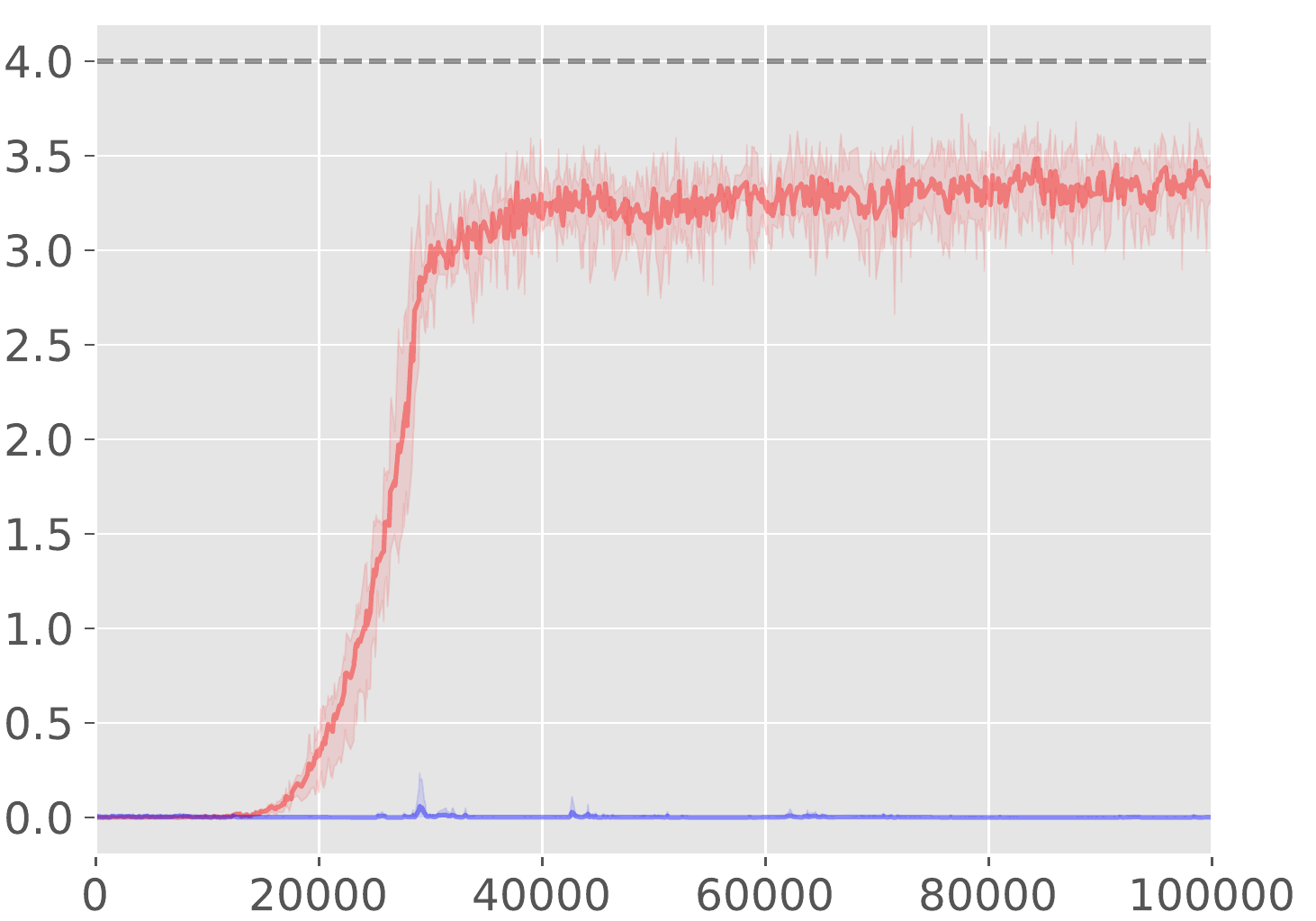} \\
\includegraphics[width=1\textwidth]{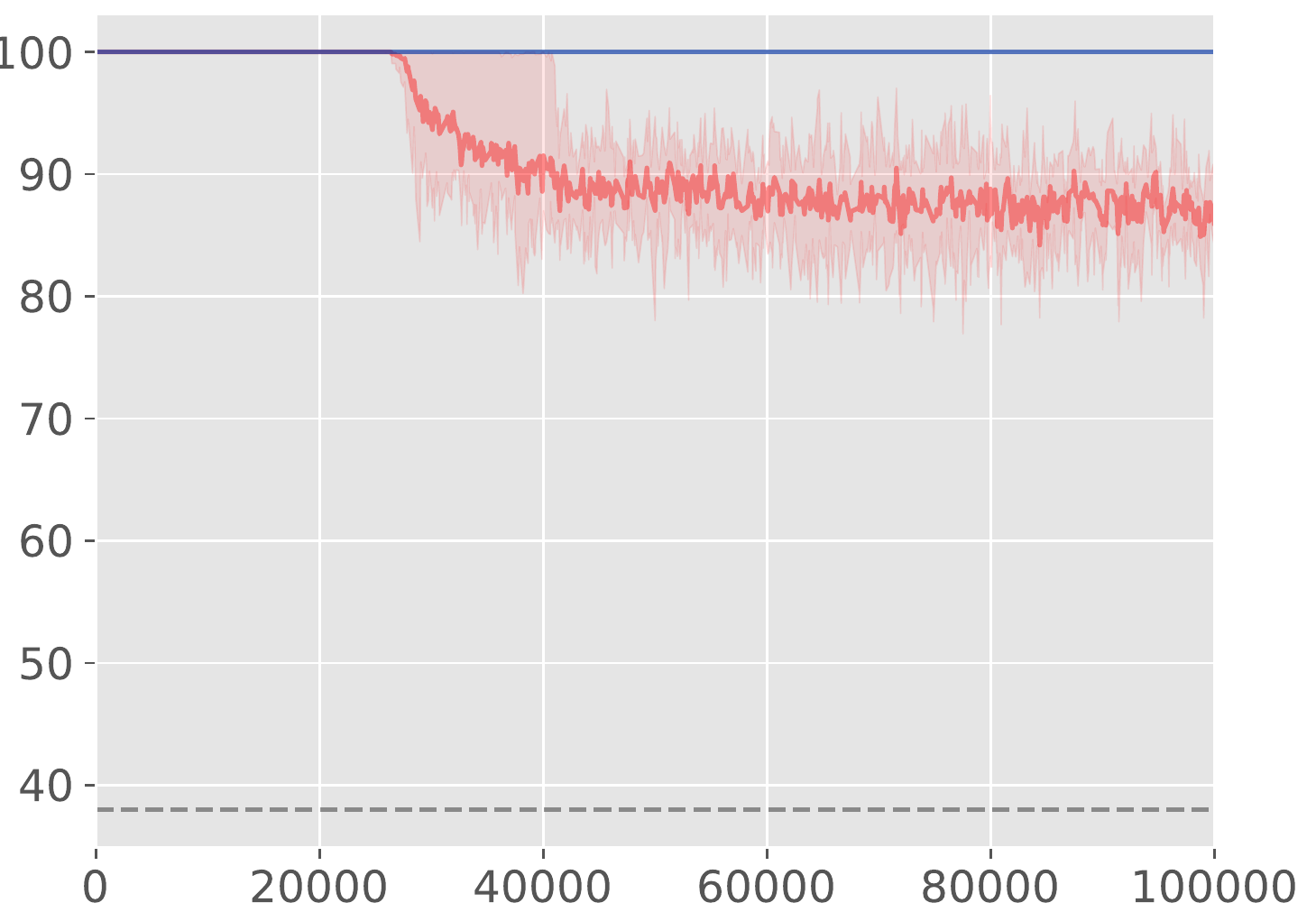}
\end{minipage}
}
\hspace{-0.2cm}
\subfigure[MATC-Coordination]{
\label{figure:coordination_results}
\begin{minipage}{0.21\textwidth}
\includegraphics[width=1\textwidth]{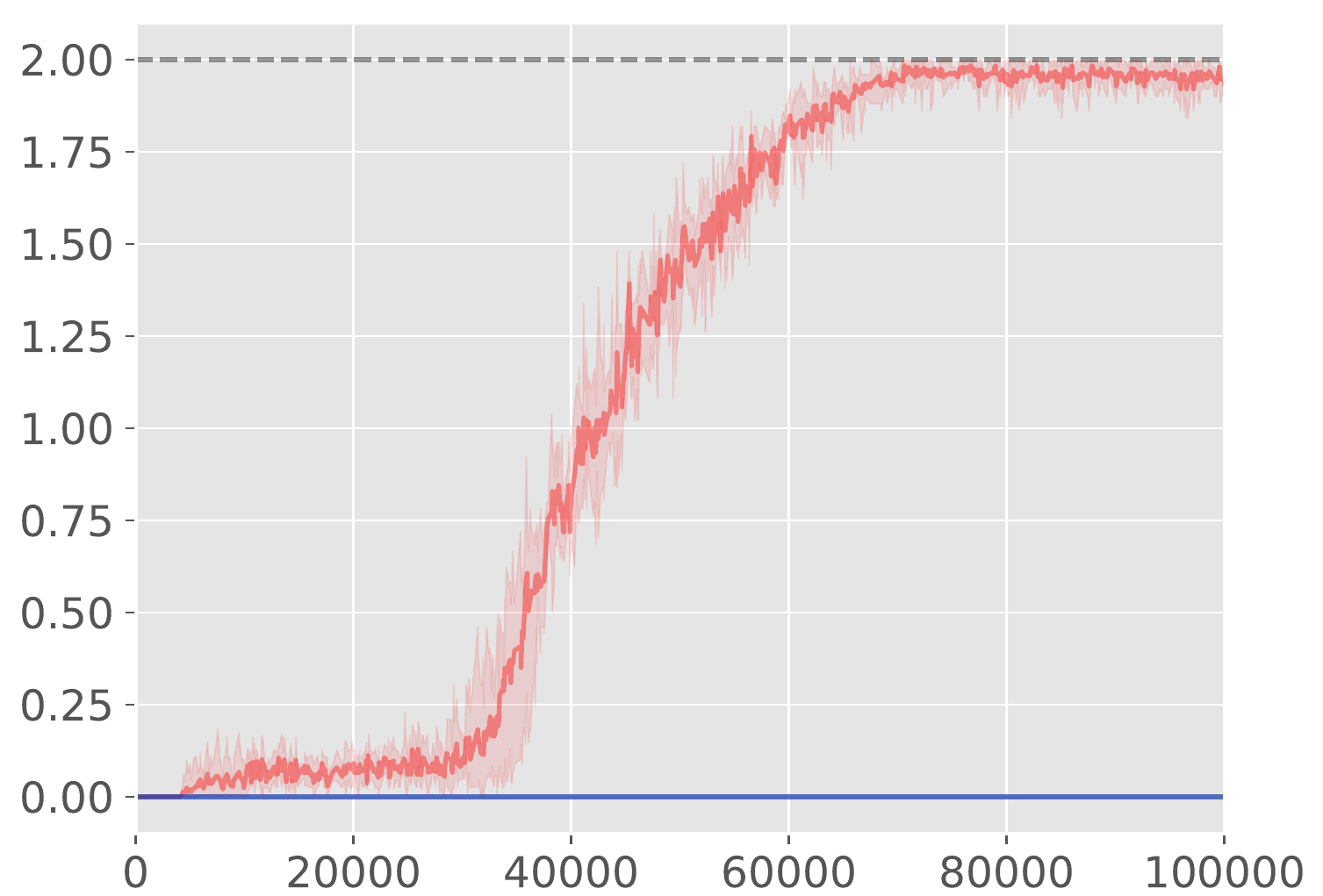} \\
\includegraphics[width=1\textwidth]{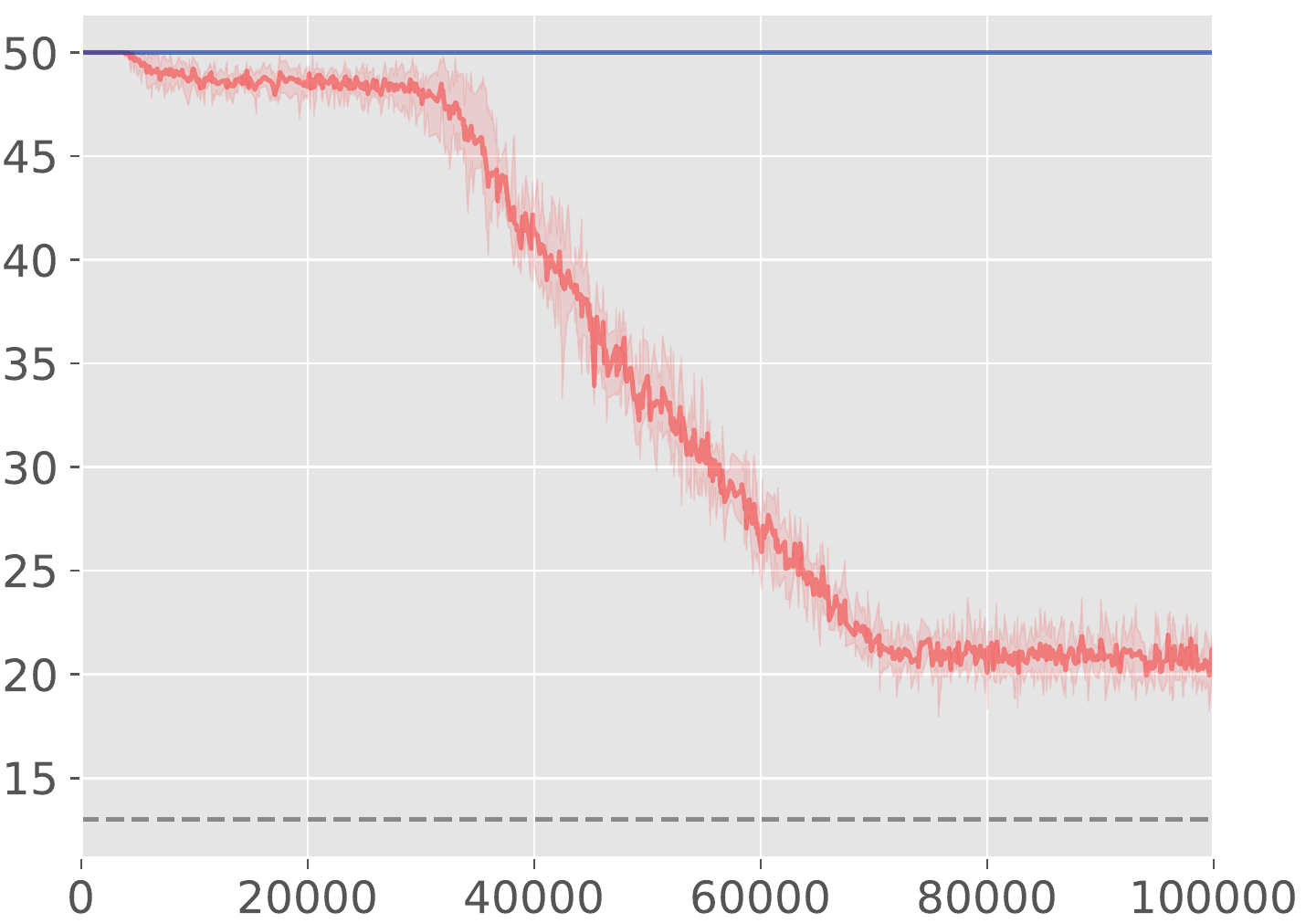}
\end{minipage}
}

\caption{
$(a)$ MATC tasks. \emph{Top-Left}: MATC-Room.
\emph{Top-Right}: MATC-Ring.
\emph{Bottom}: MATC-Coordination.
$(b)$-$(d)$ Average reward and step number of each episode over recent 100 episodes.
Horizontal axes are training episode numbers.
The results are averaged over 5 trials and the shaded region denotes a standard deviation.
}
\end{figure}

\paragraph{Temporal Abstraction}
For all three tasks, we abstract the task into two one-step operation goals (i.e., \textbf{pick-up}, \textbf{put-down}) and several navigation goals that are built over navigation actions.
For example, the navigation goals of MATC-Room are \textbf{move-to-can1}, \textbf{move-to-can2}, \textbf{move-to-trashbin1} and \textbf{move-to-trashbin2}.
Each navigation goal has a max duration of 15 time steps and can end when the goal is reached.
This induces an asynchronous termination model.
Two agents share the same task decomposition.
The low-level intrinsic observation
is a concatenation of the relevant channels, i.e., agents, walls and the target object associated with the current goal.
We use a binary intrinsic reward function for low-level policy learning, which gives 1 for reaching the goal and $-0.01$ otherwise.

\paragraph{Result Analysis}
To show the effectiveness of hierarchical deep MARL,
we compare the h-IL architecture with independent DQN (IL-DQN), which can be viewed as a non-hierarchical version of h-IL.
The results are shown in Figure \ref{figure:room_results}-\ref{figure:coordination_results}.
In all three tasks, IL-DQN can hardly achieve any positive reward, while h-IL successfully accomplishes the tasks and achieves near-optimal performance
especially in MATC-Room and MATC-Coordination.
This indicates that it is difficult to learn coordinated behaviors over primitive actions with such sparse reward.
Actually, we also apply other advanced techniques for IL-DQN (e.g., prioritized experience replay \cite{schaul2015prioritized} and SimHash exploration \cite{TangHFSCDSTA17Exploration}) while no apparent improvement is observed.
In contrast,
with hierarchical policy learning,
intrinsic goals can be easily mastered at the low level.
Further, cooperative strategies can be learned efficiently over the intrinsic goals at a large time scale.
The results for h-Comm and h-Qmix are also omitted since h-IL has already achieved near-optimal performance in such simple domain.
Experimental details are provided in the Supplementary Material.

\subsection{Fever Basketball Defense (FBD)}
\begin{figure}
\centering
\subfigure[]{
\label{figure:FBD_screenshot}
\includegraphics[width=0.52\textwidth]{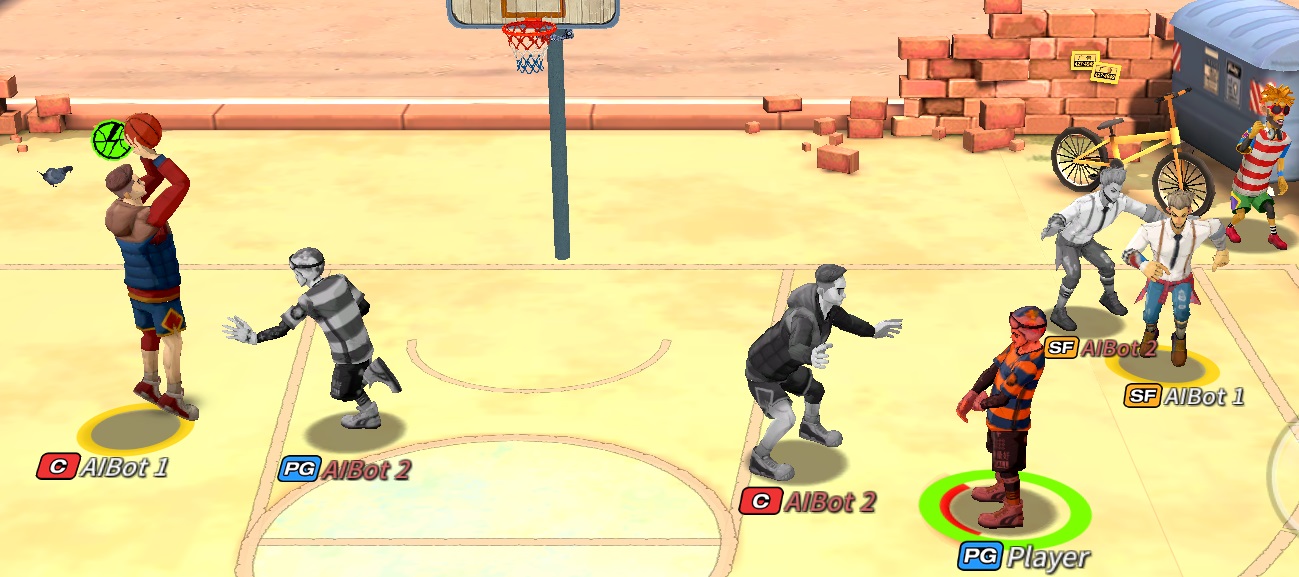}
}
\subfigure[]{
\label{figure:FBD_TA}
\includegraphics[width=0.35\textwidth]{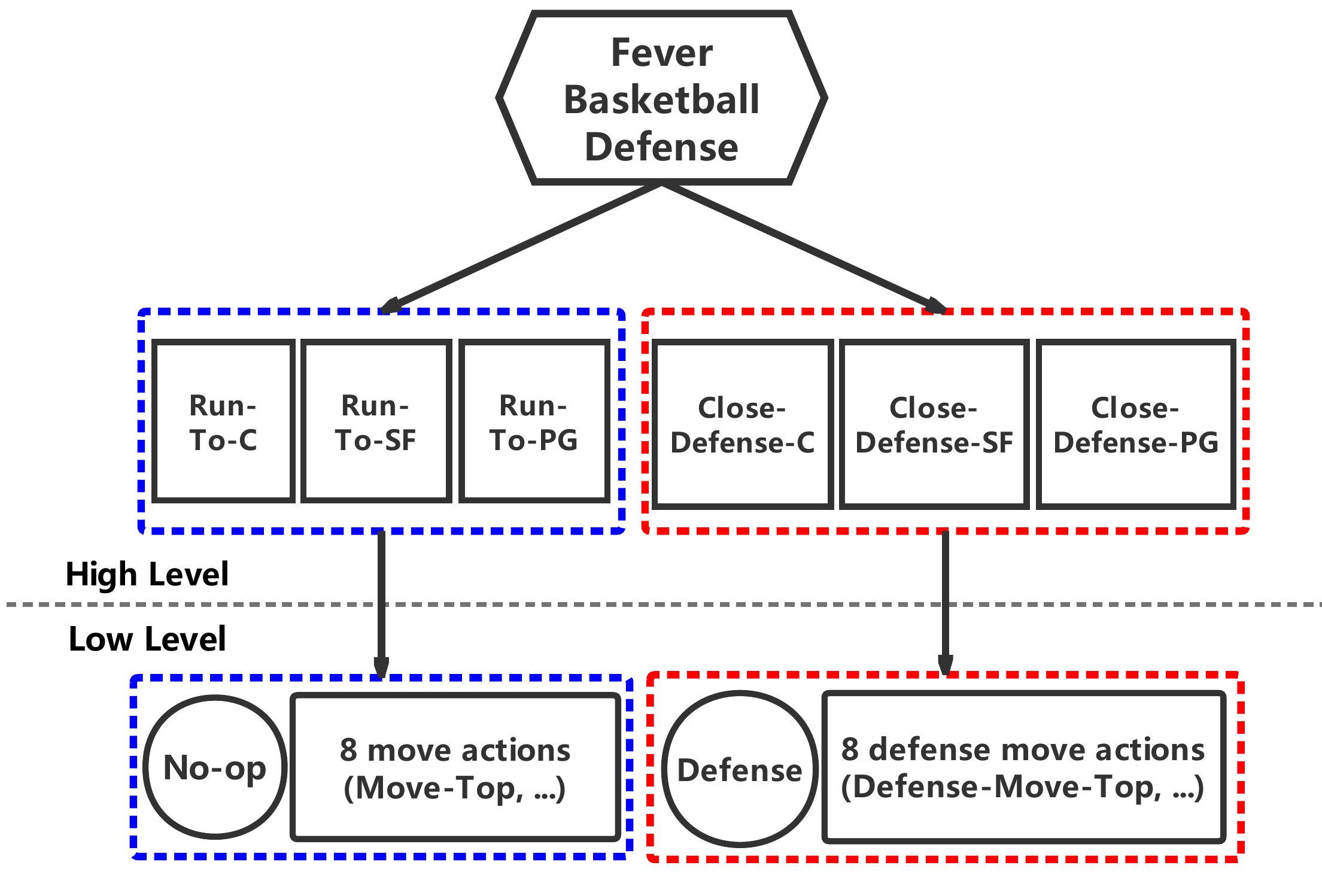}
}
\caption{Fever Basketball Defense.
$(a)$ A screenshot of Fever Basketball Defense game.
$(b)$ An illustration of the task decomposition.
The roles (i.e., C/SF/PG) in the high-level goals means the players with specific roles in the offense team.
}

\label{figure:FBD}
\end{figure}

Fever Basketball is a popular online mobile game
that mimics the real-world street basketball scenarios.
FBD is a mini-scenario of Fever Basketball
in which three defense players with different roles, i.e., Center (C),  Small Forward (SF) and Point Guard (PG), compete against the offense team.
A screenshot of FBD is shown in Figure \ref{figure:FBD_screenshot}.
The agents (defense players) aim to learn cooperative defensive strategies
to prevent the offense team from scoring.
The offense players use a built-in policy,
which is much more powerful than average human game players.

\subsubsection{Environment Setup}
Each episode renders an offense round with a max duration of 20 seconds.
A reward of $-1/+1$ is given at the end of each episode,
depending on whether the offense team scores or not.
The state is a feature vector consisting of the remaining time of the episode, the positions and types of the entities (i.e., teammates, opponents and the ball) in the game field.
Each agent has 18 available actions,
including \textbf{no-op}, 8 move actions (i.e., \textbf{move-[in 8 directions]}), and 9 defense actions (i.e., \textbf{defense} and \textbf{defense-move-[in 8 directions]}).
The defense actions, that have lower movement speed, can decrease the hit rate of the offense player nearby.
Besides, a special built-in defensive behaviors, i.e., block, can be triggered occasionally when the defense agent is in a good defense position.
Therefore, the purpose of the agents is to keep good defense positions and prevent the offense team from scoring.

\paragraph{Temporal Abstraction}
As illustrated in Figure \ref{figure:FBD_TA},
we decompose the game into a two-level hierarchy with 6 intrinsic goals (i.e., \textbf{run-to-[C/SF/PG]} and \textbf{close-defense-[C/SF/PG]}) for all three agents,
each of which has a set of 9 primitive actions.
Each goal is executed for up to 15 steps and early terminations can only happen when an episode ends or the offense players pass the ball.
This induces a synchronous termination model since agents alter their goals concurrently (the asynchronous setting is discussed in Section \ref{section:async}).
An agent's low-level intrinsic observation contains the only the information about itself and the target offense player with the specific role associated to the current goal.
To train low-level policies, we use a binary intrinsic reward function for all goals.
A reward of $+2$/$-0.01$ is given depending on whether the agent is within the Effective Defense Area, i.e., a sector area in front of the offensive agent.
For the \textbf{close-defense} goals, an extra bonus of $+2$ is given if an agent successfully blocks the ball.
Complete description can be found in the Supplementary Material.



\subsubsection{Evaluations}
We evaluate our approaches with two criteria: the miss-shot rate of the offensive team and the block-shot rate of the defensive team over recent 100 episodes.
We firstly conduct experiments with our hierarchical deep MARL architectures and two benchmark approaches, i.e., IL-DQN and Low-Level-Only.
Low-Level-Only is a variant of h-IL that trains low-level policies only and leaves high-level policy random.
We observe that CommNet and Qmix
show no better performance than IL-DQN and thus are omitted for clarity.
We suggest that the primary challenge for plain deep MARL approaches is the ineffective policy learning due to the extremely sparse reward.

\begin{figure}
\centering
\subfigure[]{
\includegraphics[width=0.445\textwidth]{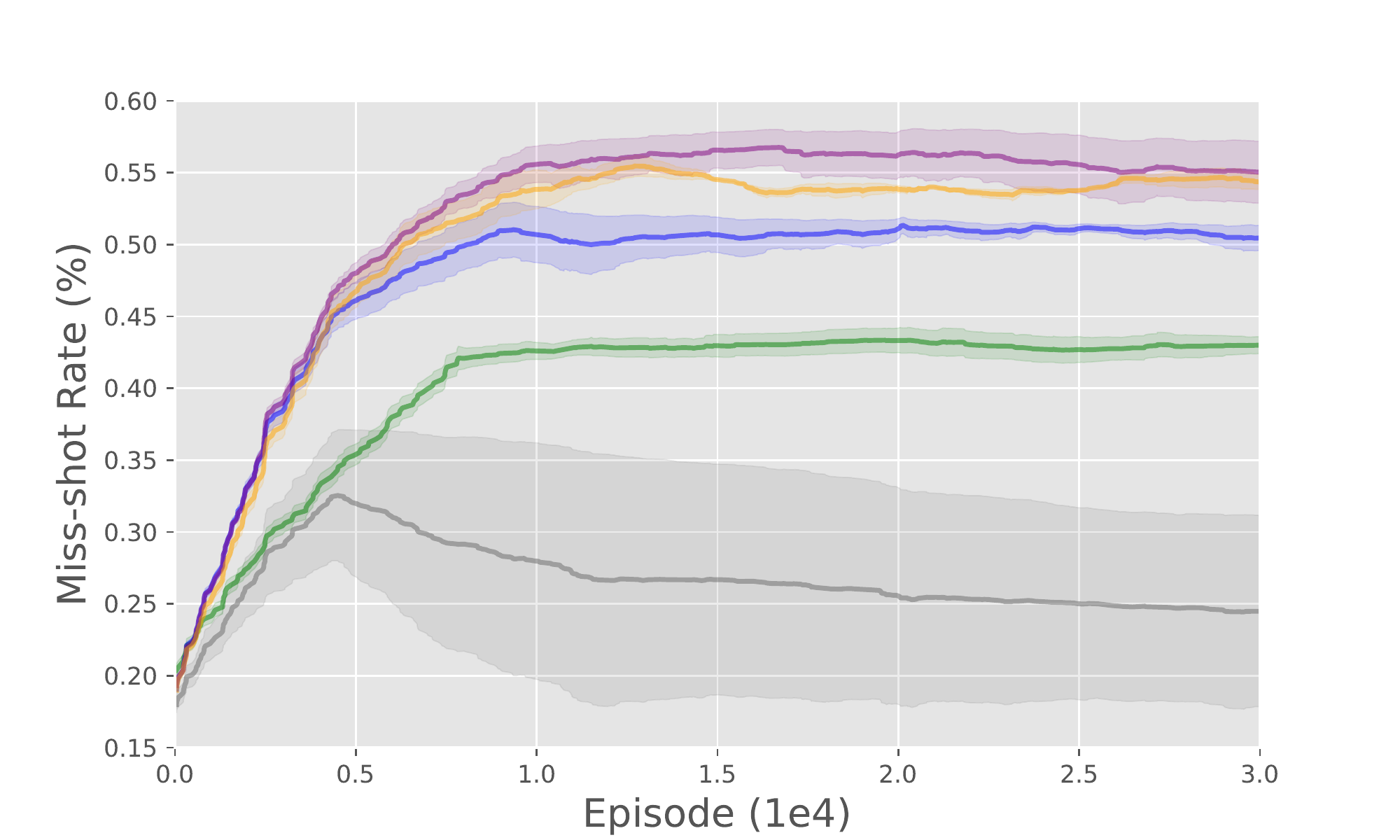}
}
\hspace{-1.25cm}
\subfigure[]{
\includegraphics[width=0.57\textwidth]{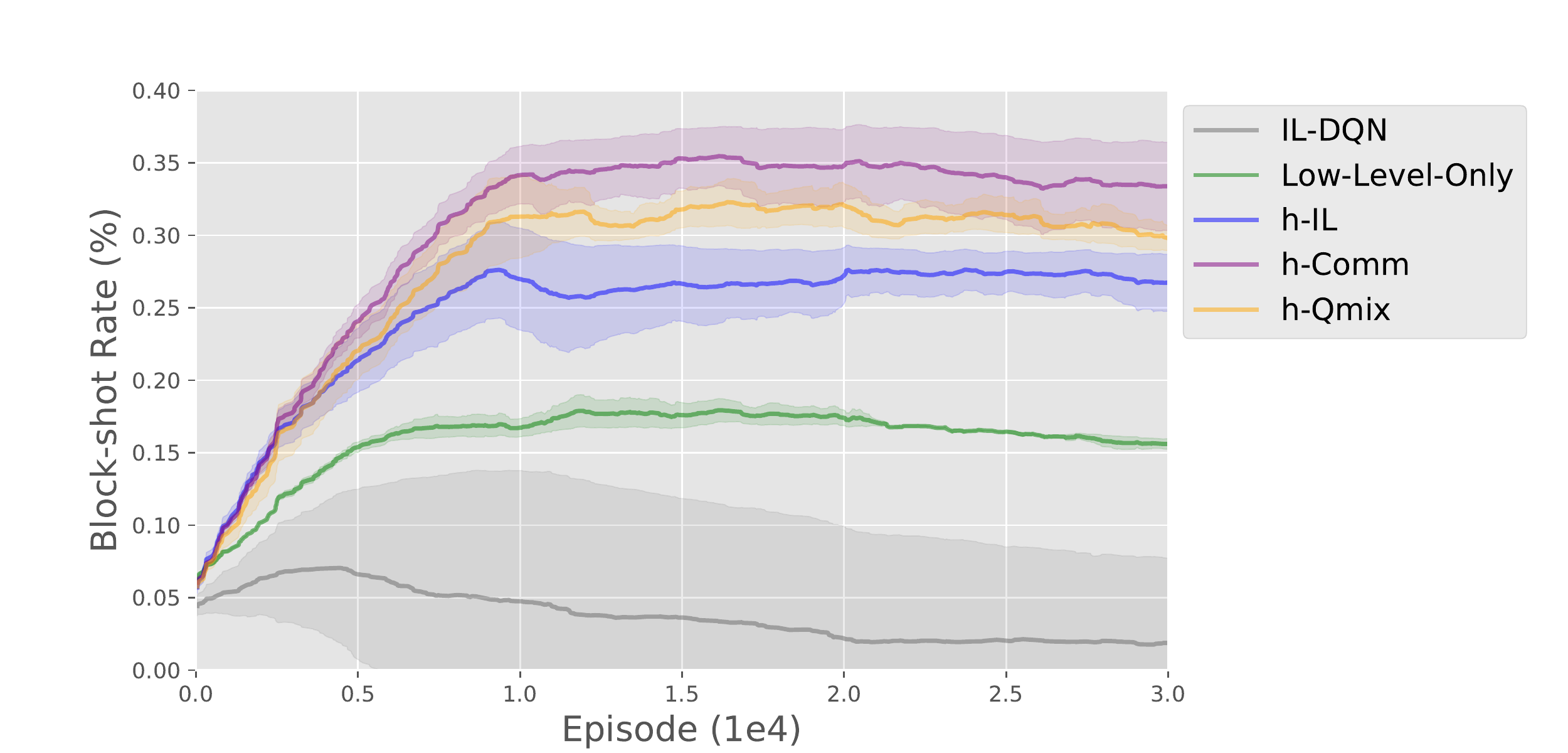}
}

\caption{
Learning curves of approaches in FBD.
The shaded region denotes a standard deviation of average evaluation over 5 trials.
}
\label{figure:FBD_results_archi}
\end{figure}

Figure \ref{figure:FBD_results_archi} shows the results for different approaches.
First,
the significance of temporal abstraction can be demonstrated by h-IL's superior performance over IL-DQN.
Besides, Low-Level-Only also shows certain defense performance even with a random high-level policy.
This is because that actions made by well-learned low-level policies still hinder the offense team.
This indicates that h-IL takes a two-step improvement via learning at different time scales.
Low-level learning is the cornerstone for effective hierarchical learning.
With high-quality low-level policies, agents can learn defense strategies efficiently at the high level, thus further elevate the defense performance.

Furthermore, h-Comm and h-Qmix achieve comparative results
and both of them clearly outperform the base architecture h-IL.
For h-Comm, we credit the advantage to the high-level communication among agents.
With explicitly taking others' information into consideration,
agents can make more informed decisions and update their policies in a more coordinated manner.
Thus, in contrast to h-IL, it is more likely for cooperative policies to emerge with the h-Comm architecture.
For h-Qmix,
a joint action-value function is trained centrally with regard to the joint rewards.
This provides individual agents with more accurate update signals
while h-IL suffers from the misleading of the joint reward.
Interestingly, we observe that h-Comm and h-Qmix show different defense tactics.
h-Comm prefers joint defense and rapid shifts,
while h-Qmix tends to one-to-one defense.
This explains the results of h-Comm and h-Qmix in Figure \ref{figure:FBD_results_archi}.
Joint defense strategy are more aggressive thus results in a higher block-shot rate, while may leave some offense player unguarded.
One-to-one defense ensures good defense quality and leads to a comparative miss-shot rate in spite of a lower block rate.
The demonstration videos can be seen in the Supplementary Material.

\begin{wraptable}{r}{0cm}
\begin{minipage}[t]{.5\linewidth}
\caption{Evaluation of ACER. The results are averaged over 5 trials.
    }
\label{table:ACER}
\centering
\scalebox{0.8}{
  \begin{tabular}{ccc}
    \toprule
    \textbf{Methods} & \textbf{Block-shot Rate} & \textbf{Miss-shot Rate} \\
    \midrule
    h-IL & $0.27$ & $0.50$ \\
    h-IL-ACER & $0.36$ & $0.55$\\
    \midrule
    h-Comm & $0.34$ & $0.55$\\
    h-Comm-ACER & $0.37$ & $0.58$\\
    \bottomrule
  \end{tabular}
}
\end{minipage}
\end{wraptable}

Next we move to the evaluation of ACER, as shown in Table \ref{table:ACER}.
The results show that ACER improves the defensive performance for both h-IL and h-Comm.
h-IL-ACER shows an significant advantage over h-IL and
achieves comparable results with h-Comm.
In contrast, h-Comm-ACER shows a slight improvement for h-Comm
and we hypothesize that it is due to the stability nature of communication and centralized learning.
This indicates that ACER is effective for stabilizing experience replay and improving high-level policy learning,
especially when agents learn their policies independently.
Note that we do not apply ACER on h-Qmix since h-Qmix naturally ensures concurrent sampling by updating policies with the joint experiences of agents.

\subsubsection{Asynchronous Termination Setting}
\label{section:async}
We also performed experiments with an asynchronous termination model by allowing agents to choose a new goal as long as it reaches/leaves the Effective Defense Area during the execution of \textbf{run-to}/\textbf{close-defense} goals.
We found similar results as the synchronous case but with a $3\%$ to $5\%$ performance decay.
We hypothesize that it is due to the non-stationary nature of asynchronous model.
Complete results are omitted due to space limitation and can be found in the Supplementary Material.

\section{Conclusion}
To the best of our knowledge,
we are the first to study hierarchical deep MARL with temporal abstraction.
In order to learn hierarchical policies efficiently with sparse rewards,
we propose three architectures, i.e., h-IL, h-Comm and h-Qmix,
and ACER.
Our experimental results show that, with our approaches,
agents can learn policies of different time scales together and 
coordination can be effectively achieved at a large time scale based on the skills mastered at the low level.
Future work will investigate the specialization for asynchronous hierarchical MARL,
the extension to competitive domains,
the automatic task decomposition of multiagent problems.

\small
\bibliographystyle{plain}
\bibliography{NIPS19-HD-MARL}

\clearpage
\appendix

\section*{A. Multiagent Trash Collection (MATC)}

\subsection*{A.1. Environment Introduction}
For all three domains in Figure 1,
the two agents (blue and red squares) are assigned to collect the cans (green squares)
and dump them into trash bin (yellow square) with impermeable walls (grey squares).
Each agent has 7 actions,
i.e., \textbf{up}, \textbf{down}, \textbf{left}, \textbf{right}, \textbf{pick-up}, \textbf{put-down} and \textbf{no-op}.
All cans and trash bins are equivalent and 7 actions are always available,
which means that agents can pick up and put down cans at any time.
Nothing happens
if an illegal action is chosen,
e.g., walk into walls, pick up or put down nothing, and pick up when currently possessing.


\begin{figure}[h]
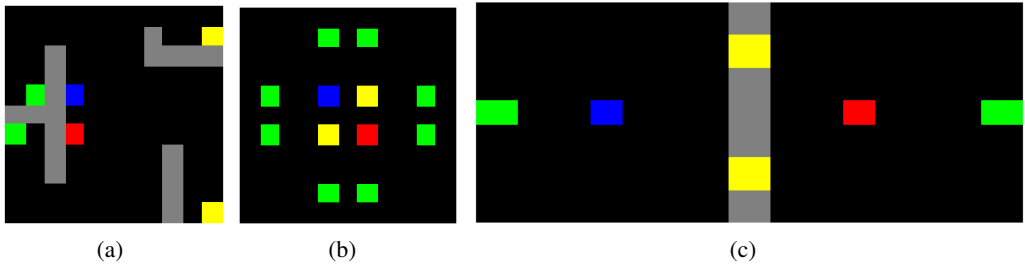

\centering
\hspace{-0.3cm}
\subfigure[]{
\includegraphics[width=0.22\textwidth]{MATC-Room-cropped.pdf}
}
\hspace{-0.3cm}
\subfigure[]{
\includegraphics[width=0.215\textwidth]{MATC-Ring-cropped.pdf}
}
\hspace{-0.3cm}
\subfigure[]{
\includegraphics[width=0.54\textwidth]{MATC-PC-cropped.pdf}
}

\caption{MATC tasks.
$(a)$ MATC-Room, 11 by 11 grids.
$(b)$ MATC-Ring, 11 by 11 grids.
$(c)$ MATC-Coordination, 15 by 7 grids.
}
\label{figure:MATC}
\end{figure}

\subsection*{A.2. Temporal Abstraction}
\begin{figure}[h]
\centering
\subfigure{
\includegraphics[width=0.9\textwidth]{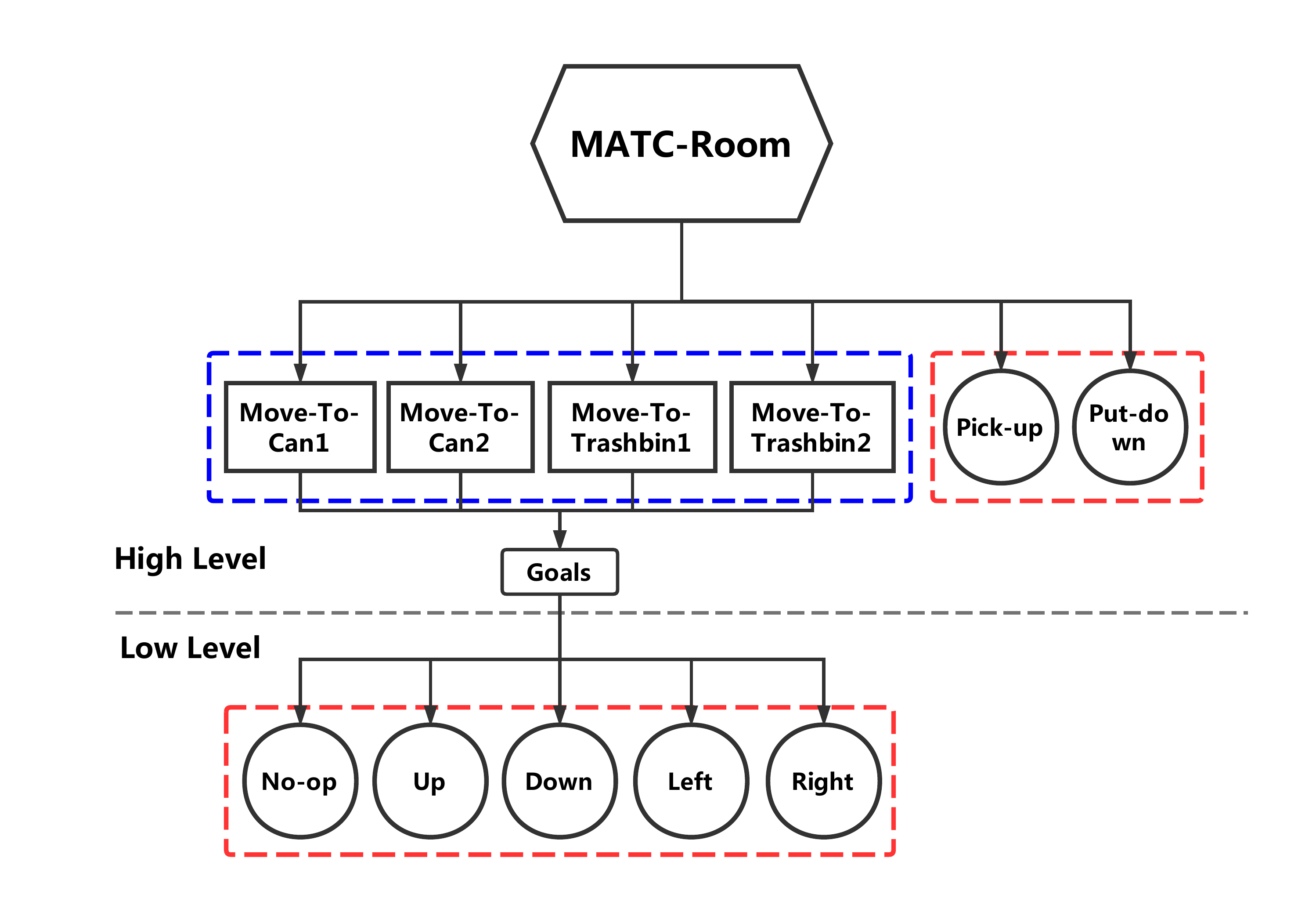}
}

\caption{The illustration of temporal abstraction for MATC-Room.
Red dashed squares denote operation goals and primitive actions (one-step execution)
and the blue dashed square is for navigation goals (multi-step execution).}
\label{figure:MATC_TA}
\end{figure}

One example of temporal abstraction for MATC-Room is shown in Figure \ref{figure:MATC_TA}.
Agents can alter their goals only when the current goal is achieved or a maximum goal length is reached.
In our experiments, we set the max goal length to be $15$ steps.
The max episode horizon is set to $100/100/50$ steps for MATC-Room, MATC-Ring and MATC-Coordination tasks respectively.

\subsection*{A.3. Experimental Details and Hyperparameters}
For h-IL, the high-level and low-level $Q$-networks are all 2-layer Multi-Layer Perceptrons (MLPs) with $64/64$ units.
The learning rates are $0.00025/0.0005$ and the discount factors are $0.95/0.9$ for high-level and low-level learning.
The memory buffer sizes are $5000/5000$ for both high level and low level.
At the beginning of training, we keep high-level policy random and update low-level policies for $50000$ times.
After the warm-up period, agents learn their high-level policies and low-level policies together.
For IL-DQN, the Q-networks are 2-layer MLPs with $128/128$ units.
The learning rates are $0.00025$ and the discount factors are $0.99$.
The memory buffer sizes are $10000$.

We conduct network updates every $20$ steps.
For h-IL, the exploration rates are annealed from $1$ to $0.1$ in $10000/100000$ updates for low-level/high-level learning.
For IL-DQN, the exploration rates are annealed from $1$ to $0.1$ in $100000$ updates.
In MATC experiments, we use parameter sharing for both high-level and low-level policies.

\section*{B. Fever Basketball Defense (FBD)}

\subsection*{B.1. Task Decomposition}
An illustration of decomposition for Fever Basketball Defense is shown in Figure \ref{figure:FBD_TA}.

\subsection*{B.2. State and Intrinsic Observation}
\begin{figure}[h]
\centering
\hspace{-0.2cm}
\subfigure{
\label{figure:FBD_TA}
\includegraphics[width=0.57\textwidth]{FBD_TA_new.pdf}
}
\hspace{-0.2cm}
\subfigure{
\label{figure:FBD_Field}
\includegraphics[width=0.4\textwidth]{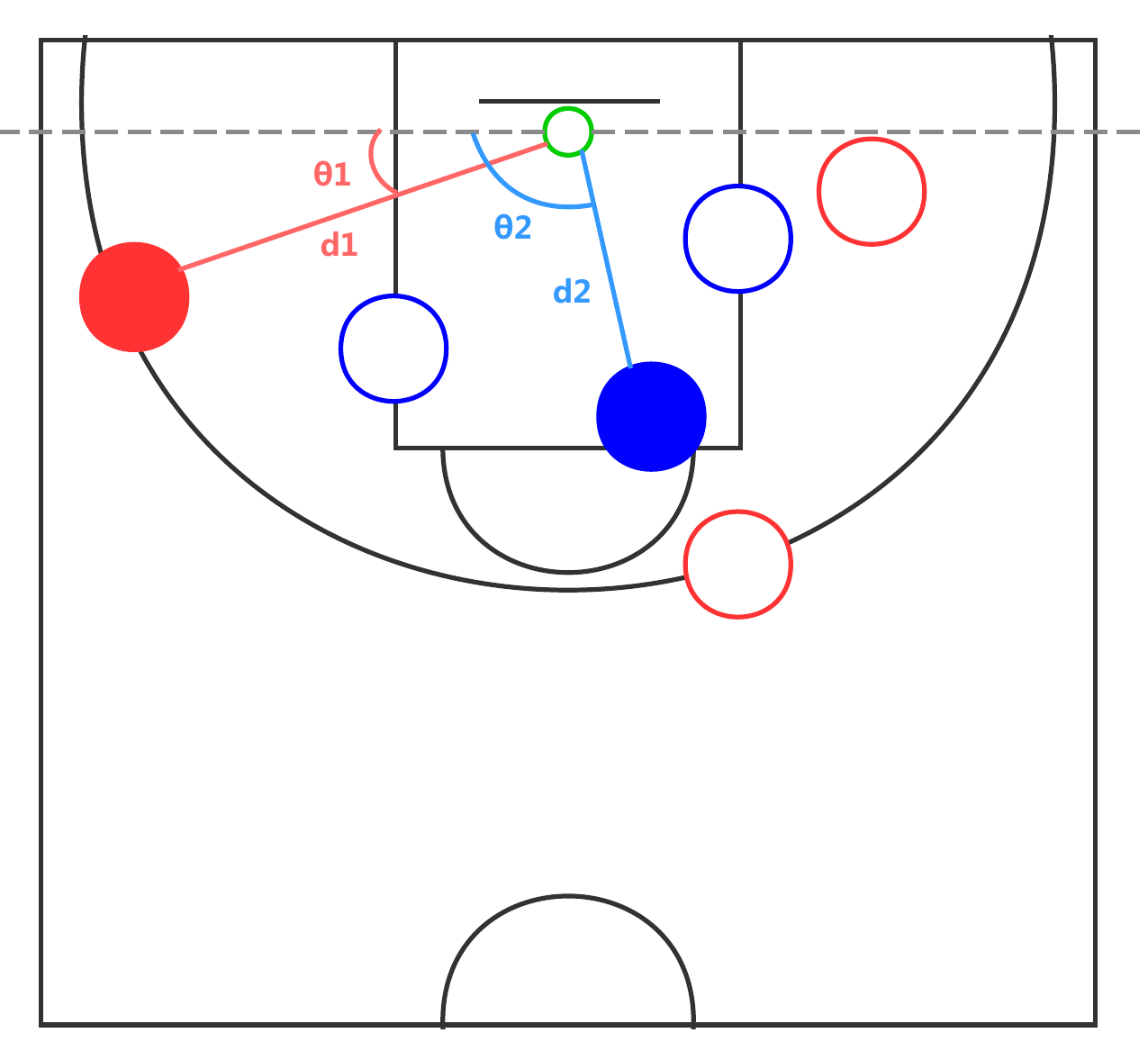}
}

\caption{
$(a)$ An illustration of temporal abstraction for FBD.
$(b)$ An illustration of the game field.
Red and blue circles denote the offensive players and the defensive players respectively.
Green circle represents the basket rim.}
\label{figure:FBD}
\end{figure}

As illustrated in Figure \ref{figure:FBD_Field},
the position of an unit, i.e., defense/offense players or the ball,
is a tuple of $\langle d, \cos(\theta), \sin(\theta) \rangle$,
where $d$ is the distance from the player to the basket rim
and $\theta$ is the angle from the horizontal axis and the line between the player and the basket rim.

The high-level learner and low-level learner receive different input features.
The state is $50$-dimension vector of features,
consisting of
the remaining time of the episode, the position of the ball,
the types and positions of itself, teammates and opponents.
With state abstraction, the intrinsic observation at the low level is a $26$-dimension vector of features
consisting of the types and positions of itself and the defense target of the current goal,
e.g., the defense target is \textbf{C} if the current goal is \textbf{run-to-C}.

\subsection*{B.3. Effective Defense Area}
The effective defense area is defined as a sector area
when an agent has a defense target.
For example, if player $2$ is the current defense target of player $1$,
we say that player $1$ is in the effective defense area
when the distance between them is less than $1.5$ meters
and the difference of their angles to the basket rim is less than $7.5$ degrees, i.e., $|\theta_1 - \theta_2| < 7.5$.

\subsection*{B.4. Experimental Details and Hyperparameters}
For h-IL,
the high-level and low-level Q-networks are all 2-layer MLPs with $128/64$ and $64/32$ units respectively.
The learning rates are $0.00025/0.00025$ and the discount factors are $0.95/0.9$ for high-level and low-level learning.
The memory buffer sizes are $20000$ for both high level and low level.
We build our h-Comm and h-Qmix architecture based on the h-IL architecture with the same hyperparameters.
For h-Comm, we implement the Comm module between two hidden layers of the high-level networks.
For h-Qmix, we use the same network design for mixing networks and hypernetworks as the original paper.

We update networks every $1/2$ steps for low-level and high-level learning respectively.
The exploration rates are annealed from $1/1$ to $0.01/0.1$ in $100000/20000$ updates
for low-level and high-level learning respectively.
In FBD experiments, we use parameter sharing for low-level networks only.
At the beginning of training, we keep high-level policy random and update low-level policies for $200000$ times.
After the warm-up period, agents learn their high-level policies and low-level policies together.

\section*{C. Additional Discussion}

\subsection*{C.1. Experience Trim for h-Qmix with Asynchronous Termination Model}
Since the high-level learning of h-Qmix follows the centralized training and decentralized execution paradigm,
the h-Qmix architecture may not be directly applicable in the asynchronous termination setting.

Concretely, consider the update process of agents' high-level $Q$-networks with h-Qmix:

\begin{equation}
\label{equation:hqmix_update}
\begin{aligned}
  Q^i(\Vec{o}_t, \Vec{g}_t) = Q^i(\Vec{o}_t, g^1_t, \dots, g^N_t) \leftarrow & (1 - \alpha) Q^i(\Vec{o}_t, g^1_t, \dots, g^N_t) \\
  & + \alpha \big[ \sum^{\tau-1}_{k=0} \gamma^{k} r^i_{t+k} + \gamma^{\tau} \max_{g^i_{t+\tau}} Q^i(\Vec{o}_{t+\tau}, g^1_{t+\tau}, \dots, g^N_{t+\tau})\big].
\end{aligned}
\end{equation}
Since multiple agents may choose and terminate their high-level decisions (i.e., intrinsic goals) at different time points,
this induces that other agents could make several intrinsic goals during the execution of $g^i_t$.

For example, agent 1 chooses an intrinsic goal at step 0, and terminates the goal at step 9.
Meanwhile, agent 2 also chooses an intrinsic goal at step 0 and terminates the goal at step 5, then chooses another goal and terminates the second goal at step 9.
Thus, there are one high-level transition for agent 1 and two high-level transitions for agent 2, resulting in an invalid update for $(\Vec{o}_0, g^1_0, g^2_0, r_{0:8}, \Vec{o}_9)$ with Equation \ref{equation:hqmix_update}.

One possible approach is the experience trim, e.g., trim the original transition of agent 1 and agent 2 into $(\Vec{o}_0, g^1_0, g^2_0, r_{0:4}, \Vec{o}_5)$ and $(\Vec{o}_5, g^1_0, g^2_5, r_{5:8}, \Vec{o}_9)$.
This aligns the experiences through cut the transition of agent 1 into two segments.
Now the update of $Q$-functions can be carried out with Equation \ref{equation:hqmix_update}.

\section*{D. Complete Results}

\begin{table}[h]
\caption{The miss-shot rates and the block-shot rates over recent 100 episodes for all approaches in Fever Basketball Defense (FBD) with synchronous and asynchronous termination models.
The results are averaged over 5 trails.
Note that h-Qmix is not applicable (N/A) in asynchronous termination setting directly.
}
\centering
\scalebox{1.0}{
\begin{tabular}{c|cc|cc}
\toprule
\multicolumn{1}{c}{} & \multicolumn{2}{c}{Synchronous} &  \multicolumn{2}{c}{Asynchronous} \\
\cmidrule(r){2-3} \cmidrule(r){4-5}
Approach & Block-shot Rate & Miss-shot Rate & Block-shot Rate & Miss-shot Rate \\
\midrule
IL-DQN & $0.02$ & $0.25$ & $0.02$ & $0.25$ \\
Low-Level-Only & $0.16$ & $0.43$ & $0.16$ & $0.43$ \\
\midrule
h-IL & $0.27$ & $0.50$ & $0.24$ & $0.48$ \\
h-Comm & $0.34$ & $0.55$ & $0.30$ & $0.53$ \\
h-Qmix & $0.30$ & $0.55$ & N/A & N/A \\
\midrule
h-IL-ACER & $0.36$ & $0.55$ & \textbf{0.31} & $0.52$ \\
h-Comm-ACER & \textbf{0.37} & \textbf{0.58} & \textbf{0.31} & \textbf{0.54} \\
\bottomrule
\end{tabular}}
\label{table:complete}
\end{table}

\section*{E. Demonstration Videos}
We provide two demonstration videos for h-Qmix and h-Comm respectively,
i.e., \textbf{h-Qmix-demo.mp4} and \textbf{h-Comm-demo.mp4}.

In the videos,
our agents always control the team that is currently doing defense,
i.e., control team A if A is doing defense and control team B when team A gets the ball and team B starts to defense.

We can observe that h-Comm and h-Qmix learns different defense strategies.
h-Qmix agents prefer to one-to-one defense and almost always keep good defense positions.
Besides, local cooperative defense can also been observed which means team spirits emerge among h-Qmix agents.
The behaviors are quite reasonable as commonly seen in real-world basketball matches.
In contrast,
h-Comm learns relatively aggressive strategies.
The agents show rapid shifts in defense target and joint defense behaviors occur a lot.
In two aspects, this improves the quality of local defense, thus resulting in a higher block-shot rate, but may cause the issue of leaving some offense player unguarded.

\end{document}